\documentclass[times,twocolumn,final,authoryear]{elsarticle}

\usepackage{prletters}
\usepackage{framed,multirow}

\usepackage{amssymb}
\usepackage{amsmath}
\usepackage{latexsym}
\usepackage{subfig}
\usepackage{float}

\usepackage{url}
\usepackage{xcolor}
\definecolor{orange}{rgb}{.8,.349,.1}
\definecolor{blue}{rgb}{.1,.1,.8}
\definecolor{green}{rgb}{0.0,0.5,0.0}
\definecolor{green1}{rgb}{0.0, 0.26, 0.15}
\definecolor{pink1}{rgb}{.87,.19,.39}
\definecolor{pink2}{rgb}{0.9, 0.17, 0.31}
\definecolor{pink}{rgb}{.8,.1,.6}
\definecolor{lavender}{rgb}{0.45, 0.31, 0.59}
\definecolor{darkpink}{rgb}{0.53, 0.15, 0.34}
\definecolor{red}{rgb}{0.55, 0.0, 0.0}
\definecolor{red1}{rgb}{0.76, 0.13, 0.28}
\definecolor{violet}{rgb}{0.54, 0.17, 0.89}

\begin{document}

\clearpage

\ifpreprint
  \setcounter{page}{1}
\else
  \setcounter{page}{1}
\fi

\begin{frontmatter}

\title{Prediction of Rainfall in Rajasthan, India using Deep and
	Wide Neural Networks}

\author[1]{Vikas \snm{Bajpai}} 
\author[1]{Anukriti \snm{Bansal}\corref{cor1}}
\cortext[cor1]{Corresponding author: 
  Tel.: +0-000-000-0000;  
  fax: +0-000-000-0000;}
\ead{anukriti1107@gmail.com}
\author[2]{Kshitiz \snm{Verma}}
\author[3]{Sanjay \snm{Agarwal}}

\address[1]{Department of Computer Science and Engineering, The LNM Institute of Information Technology, Jaipur, 302031, India}
\address[2]{Department of Computer Science, JK Lakshmipat
	University, Jaipur, 302026, India}
\address[3]{State Project Monitoring Unit, National Hydrology
	Project, Water Resources Rajasthan, Jaipur, 302017, India}

\received{1 May 2013}
\finalform{10 May 2013}
\accepted{13 May 2013}
\availableonline{15 May 2013}
\communicated{S. Sarkar}

\begin{abstract} 
Rainfall is a natural process which is of utmost importance in
	various areas including water cycle, ground water recharging,
	disaster management and economic cycle. Accurate prediction
	of rainfall intensity is a challenging task and its
	exact prediction helps in every aspect. 
	In this paper, we propose a deep and wide rainfall prediction
	model (DWRPM) and evaluate its effectiveness  to predict 
	rainfall in Indian state of Rajasthan using historical
	time-series data.
	For wide network, instead of using rainfall intensity values
	directly, we are using features obtained after applying a
	convolutional layer.
	For deep part, a multi-layer perceptron (MLP) is used.
	Information of geographical parameters (latitude and
	longitude) are included in a unique way. It
	gives the model a generalization ability, which helps a
	single model to make rainfall predictions in different
	geographical conditions.
	We compare our results with various deep-learning approaches
	like MLP, LSTM and CNN, which
	are observed to work well in sequence-based predictions.
	Experimental analysis and comparison shows the applicability
	of our proposed method for rainfall prediction in Rajasthan.
 
\end{abstract}


\end{frontmatter}


\section{Introduction}
\label{sec:introduction}
Knowledge of rainfall characteristics plays an important role in
understanding hydrology of a region as well as for efficient
engineering, planning and management of water resources
\citep{Campling98, Halbe13}.  It is one of the key natural
resources that has a varying impact on human society
\citep{le2020livelihood}, such as, agricultural activities
\citep{Bhatt13}, hydro-power generation \citep{Kumar18},
vegetation phenology \citep{Chakraborty18}, flood control
\citep{Karunasagar17, Sankaranarayanan19}, travel and maintenance
activities, and sustainability of biodiversity
\citep{Pangaluru15}.   
This rainfall prediction further helps us in estimating the water
requirement \citep{vaes2001rainfall} by the humans in a
particular
area or region. Developing countries like India have more advantage of
accurate rainfall prediction, reason being, all the developed
nations are either blessed with adequate rainfall or are equipped
with advanced irrigation, water recycling, and harvesting
facilities which ultimately makes them much more secured and
fostered in terms of water consumption, recycling and harvesting.

In India, most of the rainfall is received
during four months of monsoon season from June to September
\citep{Singh17, gadgil2003indian, kumar2006unraveling}. 
Rajasthan, a state in India comes under the arid zone \citep{goyal2004sensitivity} and rainfall is the
main source of overall water supply in the state. 
Rajasthan observes a varied climate range and had observed various
floods \citep{mishraexamination} and droughts
\citep{gadgil2005monsoon, gadgil2002forecasting, preethi2011anomalous} in past.
With a lot amount of
uncertainty in Meteorology \citep{curci2017assessing}, accurate
prediction becomes a daunting task.

Prediction of occurrence and intensity of rainfall
requires a lot of efforts right from data collection
\citep{sapsford1996data, weller1988systematic}, data
cleansing\citep{hernandez1998real}, data
analysis\citep{agresti2003categorical}, data
modelling\citep{benyon1996information} and finally estimation and
prediction with the help of a suitable model.
There are countless parameters which derive the possibility of
rainfall and its intensity over a region such as temperature of
the air, distance of an area from ocean, amount of moisture
contained by winds, distance  from the mountain or mountain
ranges, altitude of a land from sea level etc.  We propose a data
centric approach, where in place of receiving, recording and
maintaining several parameters, we are using daily rainfall data
of 71 years (from the year 1957 to 2017), obtained from
rain-gauge stations installed in 33 districts of Rajasthan by
Hydrology Department, Revenue Department and Indian
Meteorological Department (IMD). This analysis on 71 years of
data itself acts as a major contributor to the overall research
and analysis we did.

In this work, we design and compare
advance deep-learning models to identify patterns from the
historical daily rainfall data of Rajasthan. For this
purpose, we adapt and improve a wide and deep learning-based
model, originally
proposed by  Cheng \textit{et al} \citep{Cheng16} for recommender
systems. We name our
model as Deep
Wide Rainfall Prediction Model (DWRPM) and
evaluate its effectiveness in accurate rainfall prediction.
The proposed model is compared
with similar advance deep-learning-based models like multilayer
perceptron, convolutional neural network and
long-short-term-memory-based recurrent neural network.

The rest of the paper is arranged as follows.
Section~\ref{sec:relatedWork} reviews the related work.
Section~\ref{sec:DWRPM} explains the proposed Deep and wide
rainfall prediction model (DWRPM). Details of experimental
evaluations, model training and results of rainfall prediction
are discussed in Section~\ref{sec:experimentalEvaluations}.
Finally we conclude the paper in
Section~\ref{sec:Conclusion} and provide avenues for future research.

\section{Related Work}
\label{sec:relatedWork}
Rainfall prediction methods are broadly classified into following
four categories: empirical \citep{al2018estimation,
awadallah2017assessment, alhassoun2011developing}, numerical
\citep{ducrocq2002storm, calvello2008numerical}, statistical
\citep{li2010improved, montanari2008estimating} and machine
learning \citep{cramer2017extensive, xingjian2015convolutional}.
Due to non-linear nature of Indian rainfall
\citep{singh2012prediction}, machine learning-based models are
gaining more popularity over empirical, numerical and statistical
methods for accurate prediction of rainfall events\citep{Singh17}.
With more focus on artificial intelligence and availability
of high computational devices, these methods
have gained a lot amount of attention in the field of
prediction and estimation
\citep{ko2020development,shah2018rainfall, Liu19, Nayak13,
Darji15}.
Recently, machine learning and deep learning-based approaches,
such as support vector machine (SVM) \citep{OrtizGarcia14},
artificial neural networks (ANN) \citep{Acharya13, Singh13,
Sahai00}, multilayer perceptron (MLP) \citep{esteves2019rainfall}, recurrent neural networks
(RNN) \citep{ni2020streamflow} and convolutional neural networks (CNN) \citep{zhang2020surface} have
become popular for predicting rainfall intensity.  Some hybrid
models have also been proposed for the purpose of rainfall
prediction \citep{Dabhi14}. 

Many researchers have proposed different ways to predict rainfall
in Rajasthan \citep{dutta2013predicting, singh2012probability},
in India \citep{Dubey15, Zhang20} as well as in abroad
\citep{Zaman18, Kashiwao17}. 
There are methods that use multiple parameters
for rainfall forecasting \cite{zhang2020surface,pham2020development}, while some time-series-based methods
use a single parameter \citep{Sahai00, Singh17}. 

Zhang \textit{et al} \citep{zhang2020surface} designed a
high-altitude combined rainfall forecasting model that uses
convolutional neural network for rainfall prediction in next 12
hours in China.
Authors have used data from 92 meteorology stations to test their
model. Pham \textit{et al} \citep{pham2020development} have
developed and compared several advanced artificial intelligence
models namely adaptive network-based fuzzy inference system
optimized with particle swarm optimization, artificial neural
networks and, support vector machine for the prediction of daily
rainfall in Hoa Binh province of Vietnam. Meteorological
parameters used for this purpose are maximum temperature, minimum
temperature, wind speed, relative humidity and solar radiation.
Hernandez \textit{et al} \citep{hernandez2016:rainfall} proposed a deep learning approach
 combining an auto-encoder and a multilayer perceptron to predict
 the next day rainfall. The data was collected from a period of
 2002 to 2013 from a meteorological station located in Manizales,
 Columbia. The parameters used in the experiments include
 temperature, relative humidity, barometric pressure, Sun
 brightness, wind speed and wind direction.
\citep{hardwinarto2015rainfall} predicted the monthly rainfall over
the region of East-Kalimantan, Indonesia using Artificial Neural
Network. 
\citep{beheshti2016new} used Centripetal accelerated particle swarm
optimization (CAPSO)to predict the average monthly rainfall in the
next five and ten years.
Ni \textit{et al} \citep{ni2020streamflow} developed two
LSTM-based models for the streamflow and rainfall forecasting.

Gope \textit{et al} \citep{gope2016early} proposed a model to
predict heavy rainfall events, 6 to 48 hours before the
occurrence of rainfall in two cities of India, namely Mumbai and
Kolkata. They used a stacked auto-encoder for feature learning and support
vector machine and neural networks for classification of rainfall
events above a certain threshold as heavy rainfall events. 
Dubey \textit{et al} \citep{dubey2015artificial} used Artificial Neural Network for
predicting rainfall in Pondicherry, India. Multiple
parameters, such as, minimum temperature, maximum temperature, water
vapor pressure, potential evapotranspiration and crop
evapotranspiration  were used for forecasting. 
For training and testing purposes only
800 and 200 samples respectively were used.
A big-data centric approach using Artificial Neural Network on Map
reduce framework was used by \citep{namitha2015rainfall} to
predict daily rainfall prediction in India. The authors have used
temperature and rainfall data of sixty three years provided by
Indian Meteorology Department.
Saha \textit{et al} \citep{saha2020prediction} used stacked
encoder, ensemble regression tree and ensemble decision tree for
the prediction of the Indian summer monsoon. Authors have
identified new predictors for the purpose of monsoon rainfall
forecasting.
Samantaray \textit{et al} \citep{samantaray2020rainfall} worked
on prediction of monthly rainfall in Bolangir watershed of India.
Authors designed and compared different machine learning
algorithms like recurrent neural network, support vector machine
and adaptive neuro fuzzy inference system.
Pritpal Singh \citep{Singh17} has used historical time-series
data to predict Indian summer monsoon rainfall. Author has
mentioned that there is climatic variability and large
fluctuations in monsoon rainfall in different parts of India. To
handle these issues, the author has used the dataset prepared by
\citep{parthasarathy1992indian, parthasarathy1994all}
by taking the mean of all-India rainfall values by weighing each
of the sub-divisional rainfall areas (306 well-distributed
rain-gauges).

Almost nil to very little research work is done in the field of
rainfall prediction in the Indian state of Rajasthan.
\citep{singh2012probability} analyzed daily rainfall data using 
probabilistic approach for estimating the annual one day maximum rainfall in the
Jhalarapatan district of Rajasthan. 
\citep{dutta2013predicting} made an attempt to predict the agricultural
draught by predicting agricultural yield using normalized
difference vegetation index (NDVI) and standard
precipitation index (SPI). 

Although machine learning and deep learning methods have done
significantly well in the area of rainfall prediction, most of
them are developed focusing on some target sites with similar
geographical conditions \citep{pham2020development}. These models
rarely consider the rainfall prediction capability of the same
model in different geographical areas, which makes it difficult
to verify the generalization capability of such models. 
We have designed a single model, which can work in different
geographical conditions consisting of desserts, plains and
mountains. The generalization ability of the model is validated
by 158 rain-gauge stations, which are distributed all over
Rajasthan.

The main objective of the present work is to design and compare
advanced deep-learning models to identify patterns from the
historical daily rainfall data of 33 districts of Rajasthan,
which can be used for making accurate rainfall predictions.
For this
purpose, we adapt and modify wide and deep learning model
proposed by \cite{Cheng16}.  They used a wide and deep learning framework to achieve
both memorization and generalization in one model for recommender
systems. 
Many authors have used this concept in different domains like
regression analysis \citep{Kim20}, quality prediction in industrial
process analysis \citep{Ren20}, etc.
Memorization is relevant to the events which have already happened in the past.
 Generalization, on the other hand, explores new feature combinations
 that have never or rarely occurred in the past.

In this paper, we explore wide and deep neural networks for the
purpose of prediction of rainfall occurrence and intensity in
Rajasthan and propose a Deep
Wide Rainfall Prediction Model (DWRPM) and
evaluate its effectiveness in rainfall prediction.
 For wide network, instead of using rainfall intensity values directly,
 we are using features obtained after applying a convolutional layer as
 it is very effective in learning spatial dependencies in and between
 the series of data \citep{Van20}.  For deep part, multi-layer
 perceptron (MLP) is used. 
 We compare the proposed model
with similar advance deep-learning-based models like multilayer
perceptron, convolutional neural network and
long-short-term-memory-based recurrent neural network.
\begin{figure}[htb]
	\centering
	\includegraphics[scale=0.56]{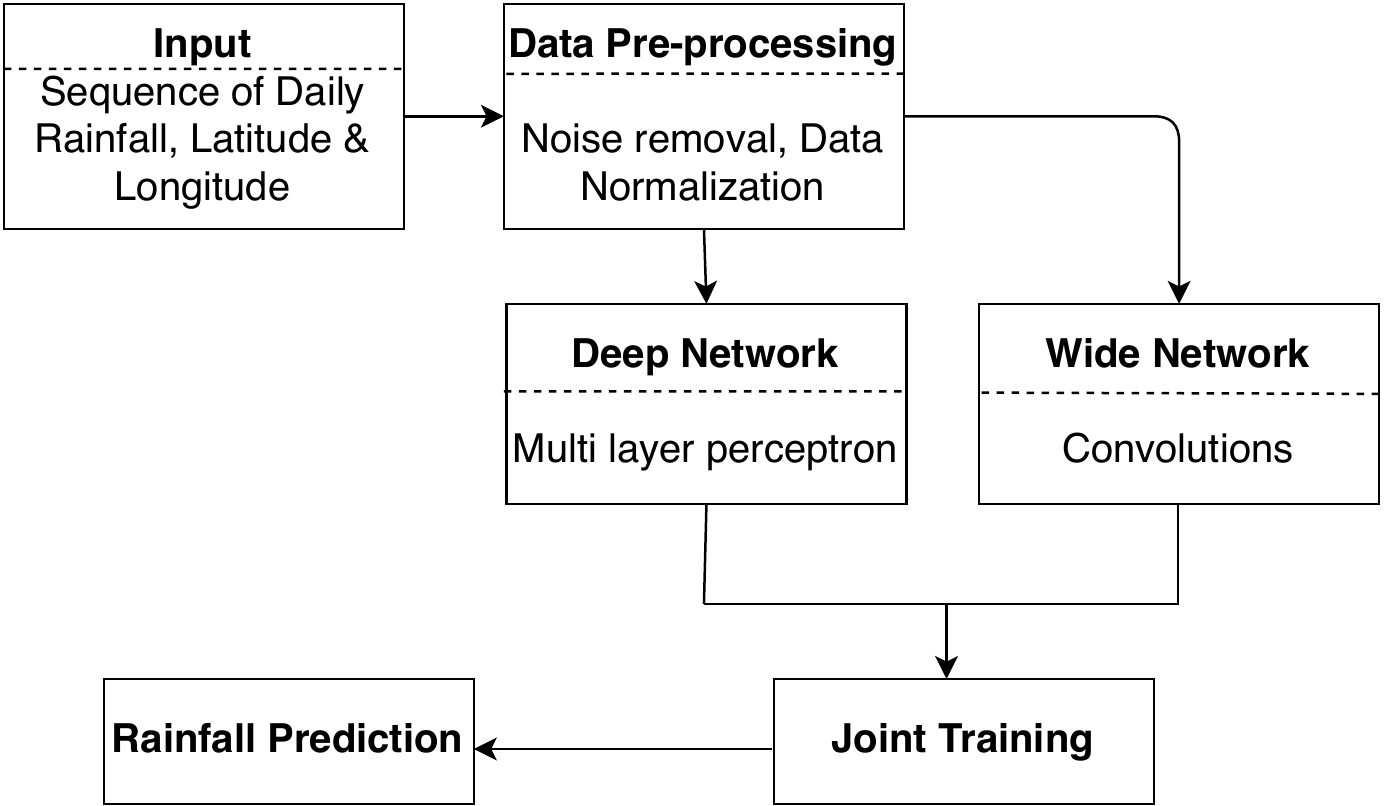}
	\caption{Overview of DWRPM: The model takes sequence of daily
	rainfall intensities and geographical parameters, namely
	latitude and longitude as input. After initial
	pre-processing, input goes to a deep network, which is a
	multilayer perceptron, and a wide network consists of
	convolutions. The model is trained using joint training
	approach, considering outputs from deep and wide networks
	simultaneously. }
	\label{fig:Overview}
\end{figure}

\section{Deep \& Wide Rainfall Prediction Model (DWRPM)}
\label{sec:DWRPM}
In this section, we first provide an overview of our approach,
and then explain various steps in the subsequent subsections.

 \begin{figure*}[!htb]
	\centering
	\includegraphics[scale=0.8]{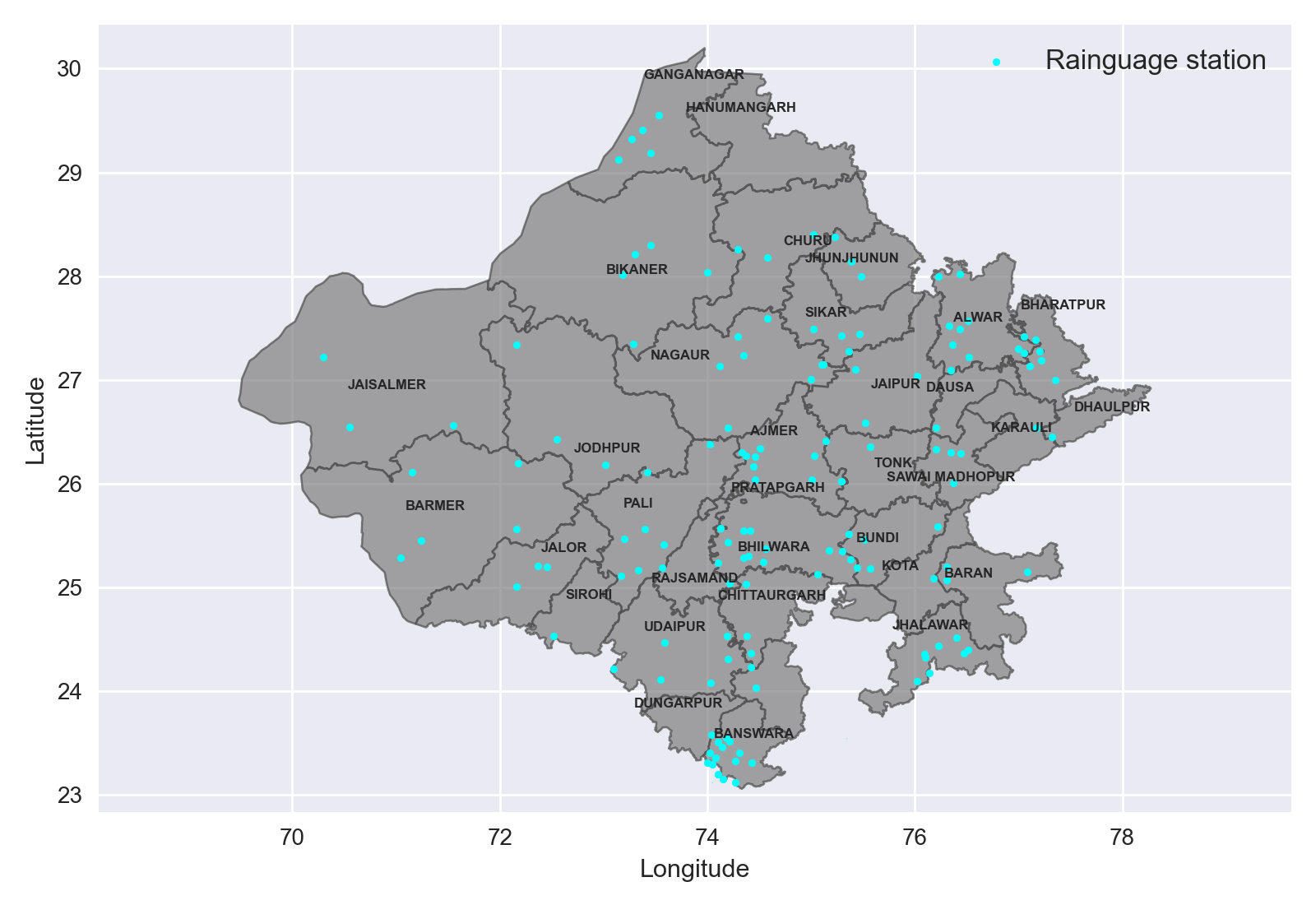}
	 \caption{Map of Rajasthan displaying the distribution of 158 rain-gauge
	 stations, shown in blue circles. Daily rainfall data of
	 these rain-gauge stations from the year 1957 to 2017 is used in the
	 training and evaluation of the proposed DWRPM.}
	\label{fig:rajasthanMap}
\end{figure*}

\subsection{Overview}
\label{subsec:overview}
The main objective of this work is to adapt and improve the Wide
and deep network for the prediction of daily rainfall using
historical time-series data of Rajasthan, India. Time
series-based prediction methods follow a concept that the current
value in a time series always depends on previous time series
values. Therefore, in this work, we use rainfall values of
previous days to predict the current day’s rainfall. For
instance, in order to predict the intensity of rainfall on
October 9, 2020, the proposed work uses previous 210 days’
rainfall intensities, i.e. from March 12, 2020, to October 8,
2020. In this work, time-series data of daily rainfall of 33
districts of Rajasthan from 158 rain-gauge stations, installed by
Hydrology Department, Rajasthan, Revenue Department, and Indian
Meteorological Department (IMD) is analyzed over a period of 71
years (from the year 1957 to 2017).  
This analysis on 71 years of
data itself acts as a major contributor to the overall research
and analysis we did. 

We are using a wide and deep neural network-based model, 
originally proposed by \cite{Cheng16} for
recommender systems. 
In our proposed work, the model
is modified and improved for the prediction of the intensity of daily
rainfall in the state of Rajasthan, India. The wide network is used to
extract low-dimensional features, using a convolutional layer.
High-dimensional features, on the other hand, are derived using
Multi-layer perceptron (MLP) \citep{pal1992multilayer} in which a 
sequence of rainfall intensity values are passed on to a deep network.
In order to incorporate a generalization ability
in the model, so that a single model can be used to make
rainfall predictions in different geographical conditions,
information of geographical parameters (latitude and longitude)
is included while training the model.
The operational steps involved in the development of our proposed
DWRPM for the prediction of rainfall are shown in
Figure~\ref{fig:Overview}.
To evaluate the performance of the
proposed method, we use two standard statistical metrics, namely
mean absolute error (MAE) and root mean square error (RMSE). We compare
our results with the advance deep learning models, widely used in
time series analysis, like MLP, one dimensional convolutional neural networks
(1-DCNN) and long short term memory (LSTM).

\subsection{Dataset description and pre-processing}
Rajasthan is the largest state of India and is situated on
the North-Western part at 27.0238$^\circ$N,
74.2179$^\circ$E. For our experiments, we have collected the daily rainfall
data of 33 districts using more than 400 rain-gauge stations, over a
period of 71 years (from the year 1957 to 2017). The dataset was
noisy, reason being several rainfall intensity
values were missing. In addition to this, some random characters were there instead of
numerical values of rainfall intensity. 
Due to administrative reasons, there were changes in total number
of districts in Rajasthan over this period of 71 years.
There were some inconsistencies in the name of rain-gauge 
stations and their coordinates which made it difficult to identity
rainfall values of a single station.
After initial pre-processing and cleansing steps, we selected 158 stations for the
purpose of our analysis. These selected rain-gauge stations are
depicted on the map of Rajasthan as shown in Figure~\ref{fig:rajasthanMap}.

	In order to provide an overview of the rainfall pattern in
	Rajasthan, an initial
	set of analysis of rainfall data obtained from the
	rain-gauge station situated at 27$^\circ$49'N,
	75$^\circ$02'E in Sikar district of Rajasthan is
	presented in Table~\ref{tab:DataStats}. We have shown
	rainfall statistics of three years, 1957, 1985 and 2017. It
	can be observed that about a significant portion of the annual
	rainfall occurs in the
	monsoon season from June to September and in the remaining
	days, there are a very few rainy day events. Daily rainfall
	generally varies
	from 0mm (no rain) to more than 500mm (in case of extremely heavy
	rainfall). Correct prediction
	of rainfall, therefore, is very important for proper
	management of water resources.

\begin{table*}[tb]
	\caption{Rainfall statistics of rain-gauge station situated at 27$^\circ$49'N, 75$^\circ$'02'E  in Sikar district of Rajasthan}
	\label{tab:DataStats}
\centering
\begin{tabular}{llllllllllllll}
	\hline
	\multirow{2}{*}{Year} & \multirow{2}{*}{Parameter} & \multicolumn{12}{c}{Months} \\
	\cline{3-14}
	 & & Jan & Feb & Mar & Apr & May & Jun & Jul & Aug & Sep & Oct & Nov & Dec \\ 
	 \hline
	 \multirow{3}{*}{1957} & Minimum Rainfall (mm) & 0 & 0 & 0 & 0 & 0 & 0 & 0 & 0 & 0 & 0 & 0 & 0  \\ 
	  & Maximum Rainfall (mm) & 15.2 & 0 & 0 & 0 & 55.9 & 77.5 & 50.8 & 54.6 & 64.8 & 11.4 & 0 & 0 \\ 
	  & Average Rainfall (mm) & 0.50 & 0 & 0 & 0 & 1.80 & 3.39 & 4.45 & 2.33 & 2.54 & 0.61 & 0 & 0 \\ 
	  \hline
	  \multirow{3}{*}{1985} & Minimum Rainfall (mm) & 0 & 0 & 0 & 0 & 0 & 0 & 0 & 0 & 0 & 0 & 0 & 0 \\ 
	  & Maximum Rainfall (mm) & 0 & 0 & 0 & 0 & 0 & 8 & 83 & 20 & 35 & 0 & 0 & 7  \\ 
	  & Average Rainfall (mm) & 0 & 0 & 0 & 0 & 0 & 0.27 & 7.61 & 2.19 & 1.43 & 0 & 0 & 0.23  \\ 
	  \hline
	  \multirow{3}{*}{2017} & Minimum Rainfall (mm) & 0 & 0 & 0 & 0 & 0 & 0 & 0 & 0 & 0 & 0 & 0 & 0 \\ 
	  & Maximum Rainfall (mm) & 16 & 0 & 0 & 0 & 12 & 24 & 32 & 14 & 7 & 0 & 3 & 1  \\ 
	  & Average Rainfall (mm) & 0.71 & 0 & 0 & 0 & 0.52 & 1.43 & 2.68 & 1.19 & 0.33 & 0 & 0.17 & 0.03  \\ 
	  \hline
\end{tabular}
\end{table*}

We have considered daily rainfall values of 210 days and geographical
parameters like latitude and longitude to predict the intensity
of next day's rainfall. 
The daily rainfall intensity ranges from 0 mm to more
	than 500 mm while latitude and
longitude ranges from 23$^\circ$12'N to 29$^\circ$55'N and
	70$^\circ$30'E to 77$^\circ$35'E, respectively. 
	Since the data is of different dimensions and dimensional
	units, therefore we normalize the data to make it dimensionally
	uniform.
When the magnitude of different parameters in a dataset is
different, the parameters with greater values often play a
major role in model training than the parameters with lower
	values. To handle this issue, we use the most commonly used min-max
normalization method to convert all rainfall intensity values to a number
	between 0 and 100 (latitude and longitude values are already
	in this range). The mathematical formula of the min-max normalization
method is as follows:

\begin{equation*}
	x^{*} = \frac{x - x_{min}}{x_{max} - x_{min}} \times 100
\end{equation*}

where, $x^{*}$ is the normalized value of the
input sample, $x$ represents a value in the original sample,
$x_{max}$ and $x_{min}$ are the maximum and minimum values, respectively.
Normalization can also help in 
	improving the learning capability of the model
	and in reducing the computational complexity
	\citep{shanker1996effect}.


\subsection{Model description}
In order to design a generalized model, which can predict
rainfall in different geographical regions of Rajasthan we design
an architecture inspired by wide \& deep networks for the
recommender systems \citep{Cheng16} and extend it for time series
based rainfall prediction. In what follows, we explain the main
components of the proposed architecture.

\subsubsection{The Wide Component: Convolutions}
\begin{figure}[htb]
	\centering
	\includegraphics[scale=0.6]{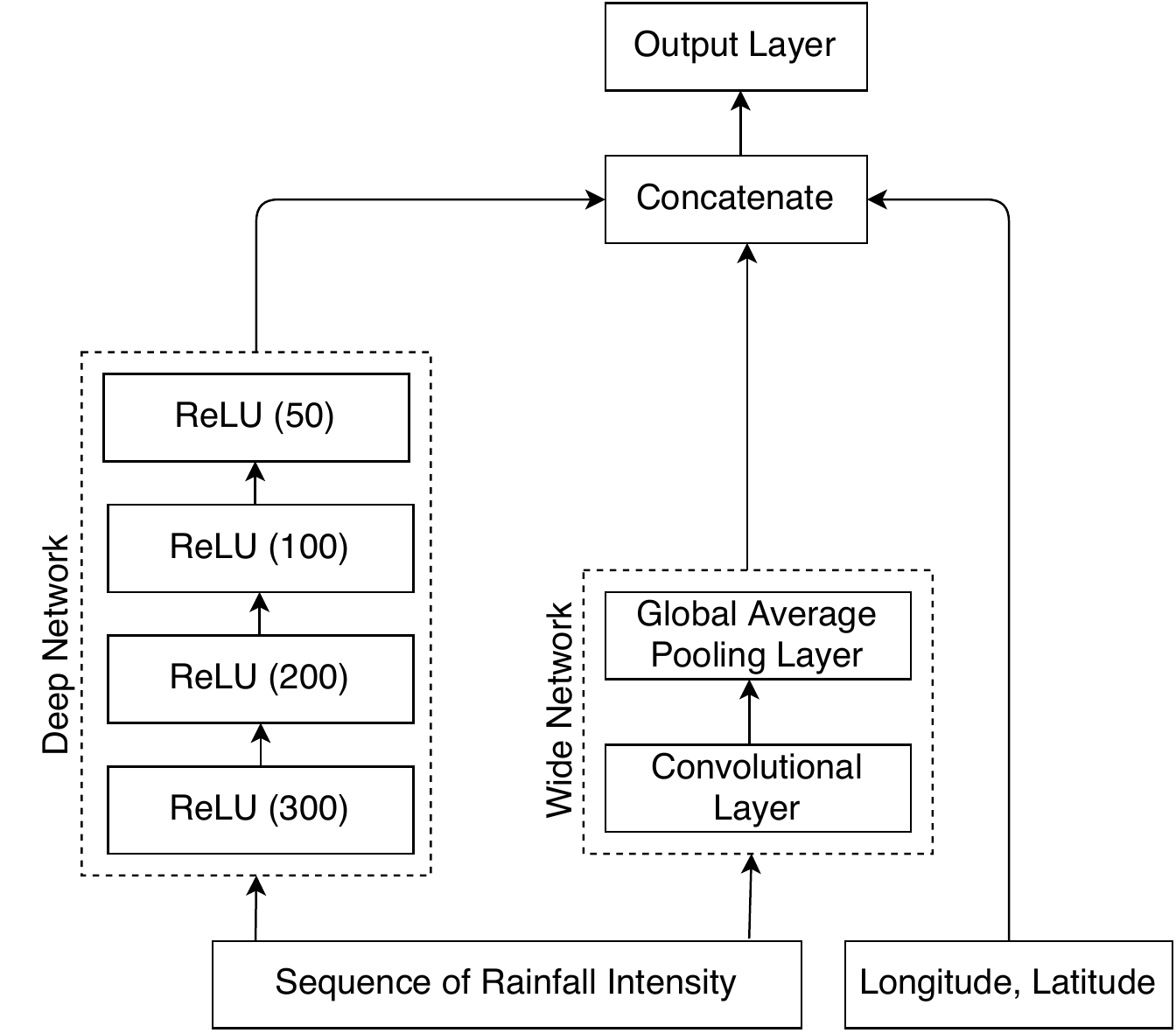}
	\caption{Selected architecture of DWRPM for rainfall
	forecasting. There are two major components: 1.The Deep
	component consists of mainly an input layer and 4 ReLU
	layers. 2. The wide component consists of a convolutional
	layer followed by a global average pooling layer. The
	information related to latitude and longitude is added
	separately. }
	\label{fig:Architecture}
\end{figure}

The wide component is used to memorize
certain combinations of rainfall events, which is beyond the
capabilities of the deep model. It is a generalized linear
model of type $y = \mathrm{\textbf{w}^{T}\textbf{x}} + b$.
In the proposed model by Cheng \textit{et al} \citep{Cheng16},
 cross-product feature
transformations were used as the wide component. 
In our proposed model 
the wide component is inspired by
convolutional neural network (CNN) as shown in Figure~\ref{fig:Architecture}.
The basic components of a general CNN consists of 2 types of
layers, namely convolutional layer and pooling layer
\citep{gu2018recent}. The convolutional layer is composed of
several convolutional kernels, which capture and learn the correlation of
spatial features by computing different feature maps. The output
of one dimensional convolutional layer with input size $N_{l}$
is:
\begin{equation*}
	a^{(l+1)}_{k} = b^{(l+1)}_{k} + \sum_{i=1}^{i=N_{l}} conv1D(w^{l}_{i,k}, a^{l}_{i})
\end{equation*}
where, $l$ is the layer number, $w^{l}_{i,k}$ is the kernel from
the $i^{th}$ neuron at layer $l-1$ to the $k^{th}$ neuron at
layer $l$, $a^{(l)}$, $b^{(l)}$ activations, bias at $l^{th}$ layer.

Convolutional layer is followed by a pooling layer that is used
to realize shift invariance by reducing the resolution of the
feature maps.
\cite{Van20} demonstrated that 1D
CNN works well in regression type
of problems and can learn the correlation in and between the
series very effectively. Therefore, instead of using raw features
in the wide part of the network, we apply a convolutional layer
to capture such combinations. 
In addition to this, to make our model more
generalized with respect to different atmospheric conditions, we
are using geographical parameters namely, longitude and
latitude while designing and developing our model (details in
Section~\ref{subsubsec:modelTraining} and Figure~\ref{fig:Architecture}).

\subsubsection{The Deep Component: Multilayer Perceptron}
The deep component is a feed-forward neural network, specifically
a multilayer perceptron, as shown in
Figure~\ref{fig:Architecture}.
Sequence of daily rainfall intensity values are given
as input, which are then fed into hidden layers of a neural
network in the forward pass. Typically, each hidden layer
computes:
\begin{equation*}
	a^{(l+1)} = f(w^{l}a^{l} + b^{l})
\end{equation*}
where, $l$ is the layer number and $f$ is the activation
function, generally rectified linear units (ReLUs), $a^{(l)}$,
$b^{(l)}$, and $w^{(l)}$ are the activations, bias and model
weights at $l^{th}$ layer.

\subsubsection{Joint training of the model}
\label{subsubsec:jointTraining}
The model is trained using the joint training approach that
optimises all parameters simultaneously by taking into
account the output of the deep and wide components, geographical
parameters and their weighted sum. 
It helps in providing
an overall prediction, which is based on aforementioned
components, also depicted in Figure~\ref{fig:Architecture}. 
\begin{equation*}
	y_{DWRPM} = \textbf{k}_{cn}\textbf{h}_{cn} +
	\textbf{k}_{co}\textbf{h}_{co} + \textbf{k}_{d}\textbf{h}_{d}
\end{equation*}
where, $y_{DWRPM}$ is the prediction, $\textbf{h}_{cn},
\textbf{h}_{co}, \textbf{h}_{d}$ are the output vectors of three
sub-models namely wide-convolutional model, wide-coordinates
model and deep model respectively, and $\textbf{k}_{cn},
\textbf{k}_{co}, \textbf{k}_{d}$ are their respective weight
vectors to be trained.

\section{Experimental evaluations}
\label{sec:experimentalEvaluations}
\subsection{Implementation details}
The experimental program is coded using Keras \citep{chollet2015keras}
API of TensorFlow framework \citep{abadi2016tensorflow, gulli2017deep}.
The computer processor is Intel
i7-8750H with 32GB RAM. Following paragraphs describe the
important aspects related to the designing and implementation of
the proposed method and the obtained results.

\subsubsection{Selection of sequence length}
As mentioned earlier, in time series-based prediction methods,
previous times series values are used to make prediction of the
next times series values. In our method, we use rainfall values
of previous 210 days as input to make predictions for rainfall
intensity of the next day. This number of days to be used as
input, is one of the hyper-parameter and its value was
chosen empirically by experimenting with different values
selected randomly in the range of 50 to 400. 
A possible reason why a sequence length of 210 gives the best
result could be, though most of the rainfall in the state of
Rajasthan is observed during four
months of monsoon i.e., from June to September, still there are some rainfall
events found to be occurring in the months of May, October and
November as well. Other days
have generally negligible rainfall intensities. There is a high
probability of capturing this correlation well when a sequence
length of 210 is considered, which is around seven months of frequent rain events.

\subsubsection{Training, validation and test sets}
\label{subsubsec:trainingTestSets}
Experimental dataset in this paper includes the rainfall data of
33 districts of Rajasthan from the year 1957 to 2017, collected
from 158 rain-gauge stations installed by
Hydrology Department of Rajasthan,
Revenue Department and Indian Meteorological Department (IMD). 
We have used data from the year 1957 to 2006 for the purpose of
training, data from the year 2007 to 2014 for validation and
finally data from the year 2015 to 2017 for
testing our model. 
This gives us 2858962 sequences for training, 429286
for validation and 140146 for testing. For creating a single
generalized model for different atmospheric conditions, we
include the geographical parameters (latitude and longitude)
of these 158 rain-gauge stations while preparing the experimental
datasets.
Indian Meteorological Department (IMD) has divided rainfall intensity
into seven categories:
\footnote{http://imdpune.gov.in/Weather/Reports/forecaster\_guide.pdf}
no rain (0mm),
light rain (0.1mm to
7.5mm), moderate rain (7.6mm to 35.5mm), rather
heavy rainfall (35.6mm to 64.4mm), heavy rain (64.5mm to
124.4mm), very heavy rain (124.5mm to 244.4mm) and extremely
heavy rain (244.5mm or more).
In our training set, we have 106452 samples of light rain, 110351
of moderate rain, 23133 of rather heavy rain, 9436 of heavy rain,
1894 of very heavy rain and only 154 samples of extremely heavy
rain events.

\subsubsection{Evaluation metrics}
\label{subsubsec:evaluation}
As shown by \cite{Glorot10} and \cite{He15}, to evaluate the overall
accuracy of predictions, we use root mean square error (RMSE) and mean
absolute error (MAE) as the basic evaluation metrics.

\begin{align}
	\label{eq:evaluationMetrics}
	RMSE &= \sqrt{\frac{1}{N}\sum^{N}_{i=1}\bigg{(}y_{i}-\overline{y}_{i}\bigg{)}^{2}} \\
	MAE &= \frac{1}{N}\sum^{N}_{i=1}\Big{|} y_{i}-\overline{y}_{i} \Big{|}^{2}
\end{align}
where, N represents the number of samples, $y_{i}$ is the
actual rainfall of the $i{th}$ sample and
$\overline{y}_{i}$ is the corresponding prediction.

\subsubsection{Model Training}
\label{subsubsec:modelTraining}
We made a parameter exploration
concerning the batch size, hidden layers, number of neurons,
dropout rates and optimization algorithms using trial-and-error
method. The network configuration of DWRPM used in our
experiments is shown in Figure~\ref{fig:Architecture}. 
The deep part is a Multi-layer perceptron with an input layer;
4 hidden layers
containing 300, 200, 100 and 50 neural units with ReLU as the
activation function; and finally a dense output layer. 
In order to prevent over-fitting of the
model, dropout layers \citep{srivastava2014dropout}
with dropout rate 0.3 are added after each hidden
layer. 
The wide part contains a convolutional layer with 100 filters,
each of size 1x5, followed by a global
average pooling layer.
The output of both the wide and deep networks is concatenated,
along with the latitude and longitude values, and the model is trained
using the joint-training approach, explained in
Section~\ref{subsubsec:jointTraining}.
We use Adam optimizer \citep{kingma14} for
training with Mean Square Error (MSE)  as loss
function. It is calculated as follows:
\begin{equation*}
		MSE = \frac{1}{N}\sum^{N}_{i=1}\bigg{(}y_{i}-\overline{y}_{i}\bigg{)}^{2}
\end{equation*}
where, N represents the number of samples, $y_{i}$ is the
actual rainfall of the $i{th}$ sample and
$\overline{y}_{i}$ is the corresponding prediction.
The goal of the model is to find optimized parameters
that minimizes MSE
\begin{equation*}
	\underset{\theta}min MSE(\theta)
\end{equation*}
where, $\theta$ is the total
number of trainable parameters.
Weights of the network are initialized using HE
initialization\citep{he2015delving}.
Model is trained for 100 epochs with batch size equals to 8.

\begin{figure*}[!ht]
\centering
	\subfloat[]{\includegraphics[scale=0.5]{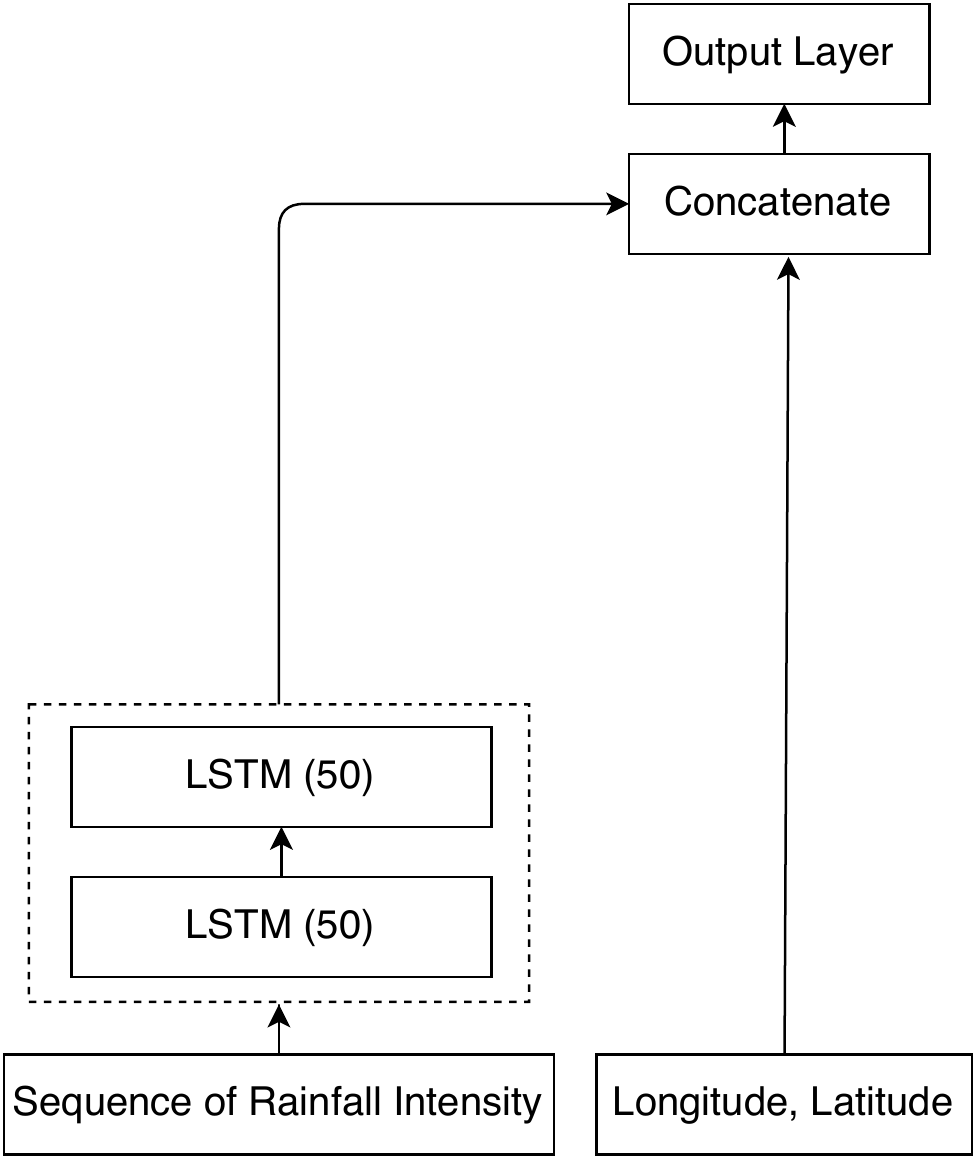}
	\label{fig:LSTMArch}} \hspace{10mm}
	\subfloat[]{\includegraphics[scale=0.5]{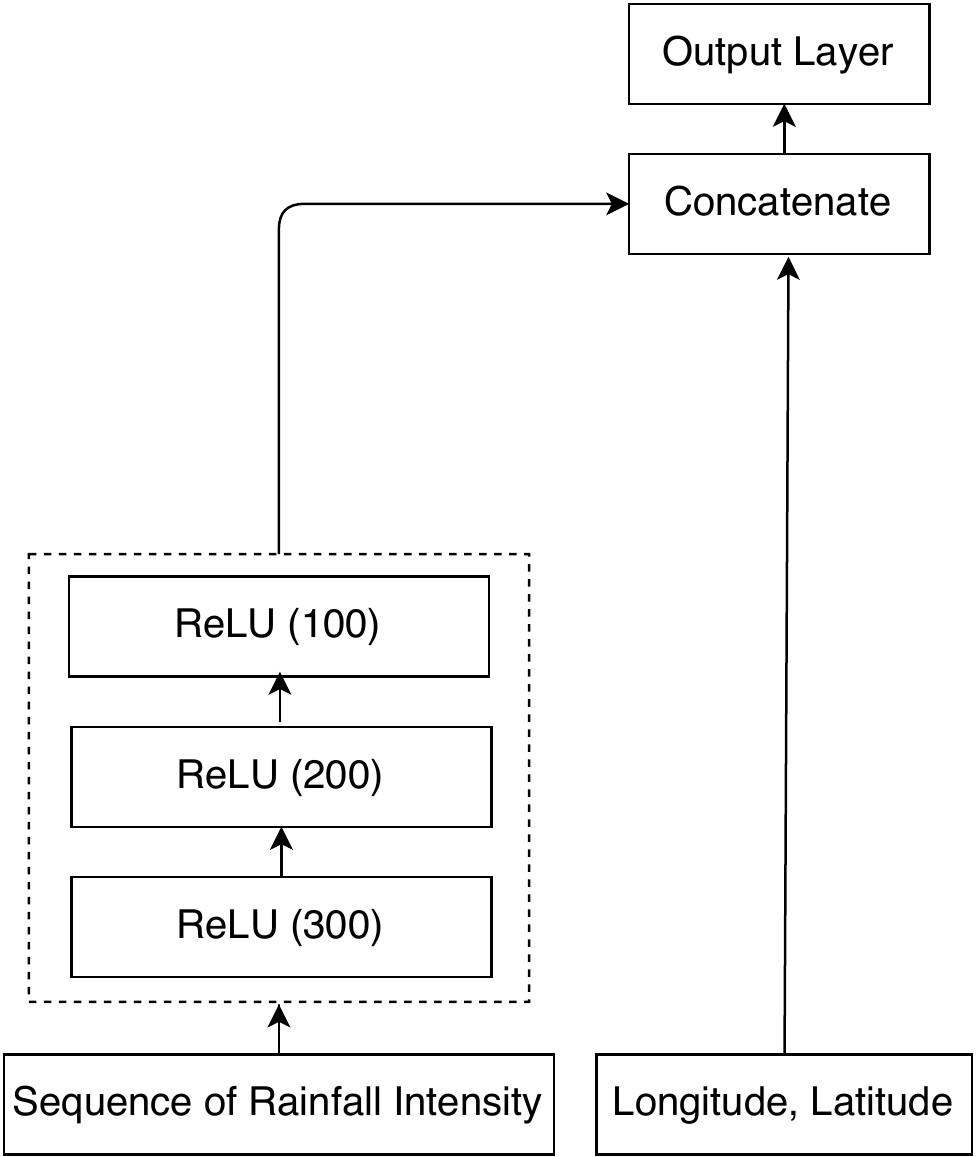}
\label{fig:MLPArch}} \hspace{10mm}
\subfloat[]{\includegraphics[scale=0.5]{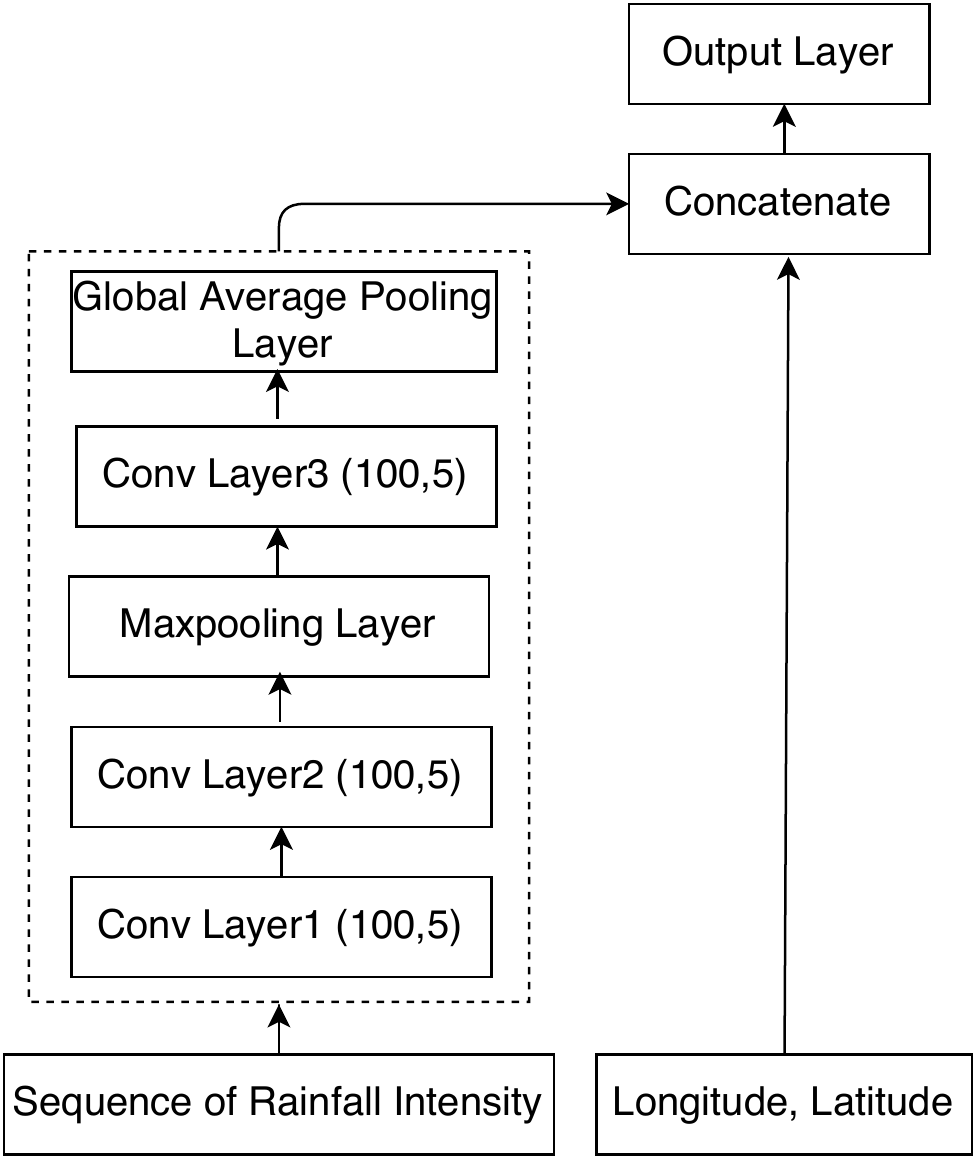}
\label{fig:CNNArch}} \hspace{10mm}
\caption{Architecture of the baseline approaches, selected after
	experimentation with various hyper-parameters. (a) Network
	configuration of LSTM (b) Network
	configuration of multilayer perceptron, and (c) Network configuration of CNN.}
\label{fig:BaselineArch}
\end{figure*}

\subsubsection{Baseline approaches}
\label{subsubsec:baselineApproaches}
To establish the efficiency of the proposed work, we compare
it with a few deep-learning-based approaches. These
approaches are observed to work considerably well for sequence-based prediction
methods (refer Section~\ref{sec:relatedWork}). We use the 
same set of input data, obtained after
pre-processing, for our proposed approach and for all the
baseline approaches.   
This is done to avoid discrepancies emerging from different sets
of input data.
The network architecture of the baseline approaches, which is
selected (after experimenting with various hyper-parameters)
for the comparative analysis with the proposed method is
explained in the subsequent paragraphs.
In all these approaches, we use Adam optimizer for training
and MSE as loss function. Input sequence length is 210.
Besides this, we concatenate the latitudinal and
longitudinal values separately, so that the sequence of rainfall intensity values 
can be learned efficiently and  overall
learning process gets enhanced (Figure~\ref{fig:BaselineArch}).

\begin{itemize}[]
	\item[] \textbf{Long Short-term Memory (LSTM):}
		The network architecture for LSTM is shown in
		Figure~\ref{fig:LSTMArch}.
		We found that this sequence
		network works well with two LSTM cells, each of size 50. 
		The output of the second LSTM layer is combined with
		coordinate values, which is finally provided to an 
		output layer for
		predicting the value of rainfall intensity. 		
		Here as well, to prevent
		the over-fitting, we use an intermediate dropout layer at
		the rate of 0.3. 

	\item[] \textbf{Multilayer perceptron (MLP):} 
		The network architecture for MLP is shown in
		Figure~\ref{fig:MLPArch}. It contains an input layer and 3
		hidden ReLU layers with 300, 200 and 100 neural units
		respectively.
		The output of the last hidden layer is concatenated with
		latitude and longitude values and fed into the dense
		layer. Dropout layers at rate 0.3 are added after each hidden layer,
		as a regularization method to prevent over-fitting.

	\item[] \textbf{Convolutional Neural Network (CNN):} The network
		architecture selected for CNN is given in
		Figure~\ref{fig:CNNArch}.
		It has two convolutional layers with 100 filters of size
		1x5
		each, followed by a max pooling layer, convolutional layer and a global
		average pooling layer. Coordinate values are appended
		with the output of global average pooling
		layer, which is finally given to the output layer for
		prediction. Dropout method is used as a
		regularization technique to handle over-fitting. The
		dropout rate is chosen as 0.2. 

\end{itemize}

\subsection{Results and discussion}
\label{sec:results}
In the following subsections, we present the results of
experimental analysis and comparison of the proposed method with
the baseline approaches described in Section~\ref{subsubsec:baselineApproaches}.

\subsubsection{Forecasting accuracy of DWRPM}
\begin{figure*}[!ht]
\centering
	\subfloat[]{\includegraphics[scale=0.55]{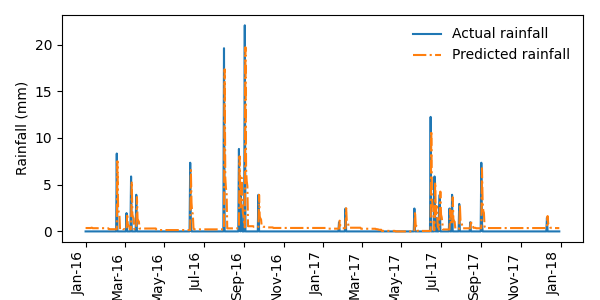}
	\label{fig:Z1}} 
	\subfloat[]{\includegraphics[scale=0.55]{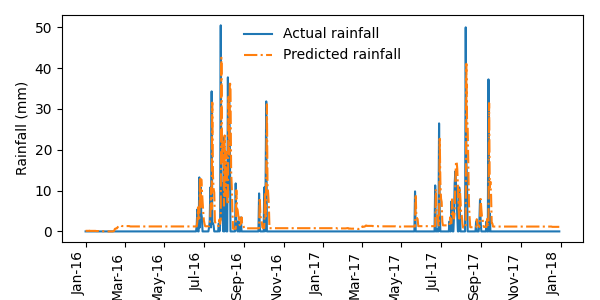}
	\label{fig:Z2}}	\\
	\subfloat[]{\includegraphics[scale=0.55]{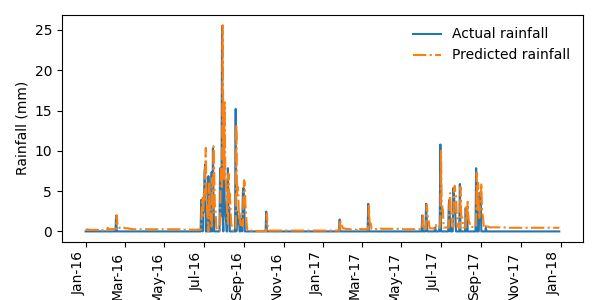}
	\label{fig:Z3}} 
	\subfloat[]{\includegraphics[scale=0.55]{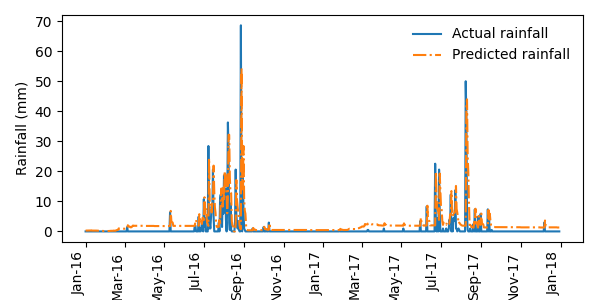}
	\label{fig:Z4}}
\caption{Prediction results of DWRPM for four rain-gauge
	stations, each picked from a different atmospheric zone
	(Section~\ref{subsubsec:generalization}). Results are from
	year 2016 to 2017. (a) Prediction
	results of rain-gauge station situated at 29$^\circ$12'N,
	73$^\circ$14'E in North-West dessert region, (b) Prediction
	results of rain-gauge station situated at 26$^\circ$04'N,
	74$^\circ$46'E in Central Aravalli hill region, (c) Prediction
	results of rain-gauge station situated at 26$^\circ$41'N,
	75$^\circ$14'E in Eastern plains region, and (d) Prediction
	results of rain-gauge station situated at 25$^\circ$18'N,
	75$^\circ$57'E in South-Eastern plateau region.}
\label{fig:Results}
\end{figure*}

Before testing the proposed model for rainfall forecasting in 158
rain-gauge stations, we need to check its stability
and feasibility. For this purpose, we take rain-gauge station
situated at degrees as an example to check the efficiency of model
in prediction of rainfall. Figure~\ref{fig:Z1} shows its prediction results
from the year 2016-2017.
Once the model is stabilized for this rain-gauge station, by tuning various
hyper-parameters (details of hyper-parameters and model training
are given in Section~\ref{subsubsec:modelTraining}), it is used for prediction of daily
rainfall intensity values for all 158 stations.
Out of total 140146 samples of test dataset, observed between the year 2015-2017,
proposed model is found to be more accurate in the prediction of light rainfall and moderate
rainfall events (0.1mm-35.5mm).
However, the accuracy of the model is comparatively lesser for very
heavy rainfall events. The main reason is that the
number of heavy
rainfall events are significantly lower than the light and
moderate rainfall events (refer
Section~\ref{subsubsec:trainingTestSets}). In addition to this,
we  are only using the
rainfall intensity values as a major parameter for prediction,
therefore it is difficult to
learn such characteristics of rainfall. Besides this, too less rainfall may
not activate neurons. 

\subsubsection{Generalization ability of DWRPM}
\label{subsubsec:generalization}
Atmospherically, Rajasthan is divided into four zones, namely: 
North West Desert Region, Central Aravalli Hill Region, 
Eastern Plains and South Eastern Plateau Region \citep{Upadhyaya14}. 
Details of the
districts, which come under the respective zones are given below:
\begin{itemize}[]
	\item[] \textbf{North-West Desert Region:} Jaisalmer, Jodhpur, Hanumangarh, Shriganganagar, Barmer, Churu, Nagaur, Pali, Sikar, Bikaner and Jhunjhunu 
	\item[] \textbf{Central Aravalli Hill Region:} Udaipur, Dungarpur, Sirohi, Jalore, Pali, Banswara, Bhilwara, Chittorgarh, Rajsamand and Ajmer
	\item[] \textbf{Eastern Plains:} Alwar, Bharatpur, Tonk, Sawai Madhopur, Karauli, Jaipur, Dausa and Dhoulpur
	\item[] \textbf{South-Eastern Plateau Region:} Kota, Bundi, Jhalawar and Baran 
\end{itemize}
%

All these zones have different atmospheric and climatic
	conditions. 
	In order to verify generalization
	ability of our model, we use it for rainfall prediction in
	each zone separately.  
	The prediction results for each zone,
	on the
	basis of two evaluation criteria i.e., MAE and RMSE
	(Section~\ref{subsubsec:evaluation}) 
	are shown in Table~\ref{tab:ZoneResults}. 
	\begin{table}[htb]
	\caption{Zone-wise prediction results}
	\label{tab:ZoneResults}
\centering
	\begin{tabular}{lll}
	\hline
	\textbf{Zone Name} & \textbf{MAE} & \textbf{RMSE} \\
	\hline
	North-West & 0.679 & 1.512 \\
	\hline
	Central Aravalli Hill Region &  0.821  & 2.43 \\
	\hline
	Eastern Plains & 0.748 & 1.634\\
	\hline
	South-Eastern Platue Region & 0.903 & 2.613\\
	\hline
	Rajasthan Region & 0.776 & 2.171\\
	\hline
\end{tabular}
\end{table}

	It can be observed that
	a single model is working well in rainfall forecasting for
	different geographical conditions ranging from plains and
	plateaus  to
	desserts and hills.  	
	Overall performance of
	the model on all rain-gauge stations of
	Rajasthan
	is also mentioned in the
	Table~\ref{tab:ZoneResults} with zone name as `Rajasthan
	Region'. 
Figure~\ref{fig:Results} shows the prediction results of four
	rain gauge stations, which are randomly picked from a
	different atmospheric zone.
For each rain-gauge station, we
	have total 887 test samples from the year 2015 to 2017. The MAE
	and RMSE values of each station are given in
	Table~\ref{tab:stationResults}.
\begin{table*}[htb]
	\caption{Prediction results of four rain-gauge stations, one
	from each atmospheric zone}
	\label{tab:stationResults}
\centering
	\begin{tabular}{lllllll}
	\hline
	\textbf{Zone Name} & \textbf{District Name} & \textbf{Station
	Name} & \textbf{Latitude} & \textbf{Longitude} & \textbf{MAE}
	& \textbf{RMSE} \\
	\hline
	North-West Dessert & Ganganagar & Anupgarh & 29$^\circ$12'N &
	73$^\circ$14'E & 0.43 & 0.60 \\
	\hline
	Central Aravalli Hill Region & Ajmer & Bhinai &
	26$^\circ$04'N
	& 74$^\circ$46'E & 0.62  & 1.12 \\
	\hline
	Eastern Plains & Jaipur & Dudu & 26$^\circ$41'N &
	75$^\circ$14'E & 0.55 & 1.18\\
	\hline
	South-Eastern Plateau Region & Bundi & Patan & 25$^\circ$18'N
	& 75$^\circ$57'E & 0.67 & 1.59\\
	\hline
\end{tabular}
\end{table*}

\begin{figure*}[!ht]
\centering
	\subfloat[]{\includegraphics[scale=0.55]{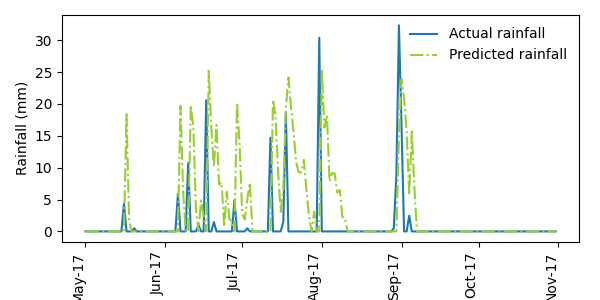}
	\label{fig:A1}} 
	\subfloat[]{\includegraphics[scale=0.55]{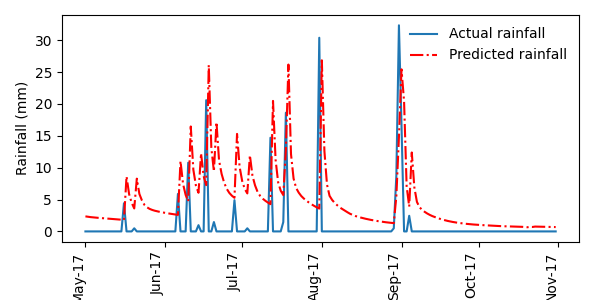}
	\label{fig:A2}}\\
	\subfloat[]{\includegraphics[scale=0.55]{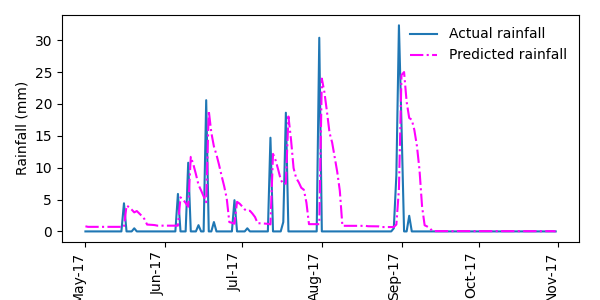}
	\label{fig:A3}} 
	\subfloat[]{\includegraphics[scale=0.55]{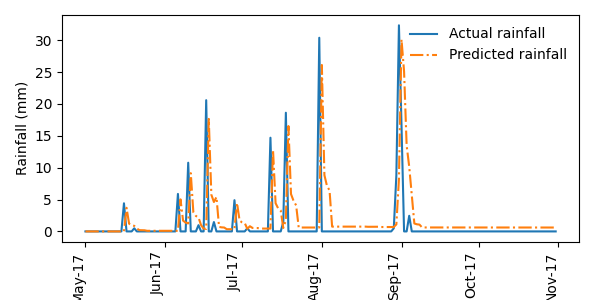}
	\label{fig:A4}}
\caption{Comparison of DWRPM and three deep-learning approaches.
Here we show prediction results of all the models on a
	rain-gauge station situated at 29$^\circ$32'N, 73$^\circ$27'E
	from May to November, of the year 2017. (a) Prediction
	results of MLP, (b) Prediction results of LSTM, (c)
	Prediction results of one dimensional CNN and, (d) Prediction results of the
	proposed DWRPM.}
\label{fig:Comparison}
\end{figure*}

\subsubsection{Comparison with baseline approaches}
To establish the significance of present work, we
compare the results of our model with the baseline
	approaches.
	Figure~\ref{fig:Comparison}
	shows prediction results of DWRPM and other approaches on a
	rain-gauge station situated at 29$^\circ$32'N, 73$^\circ$27'E
	for six months, from May to November, of the
	year 2017.
	Overall comparison of our model and other three approaches in
	rainfall prediction on all 158 rain-gauge stations from the
	year 2015 to 2017 is presented in Table~\ref{tab:Comparison}.
	It shows that the RMSE and MAE values of the proposed
	DWRPM is minimum and it gives better prediction results than
	the other
	advanced deep-learning methods, which are generally used for
	sequence-based predictions. The main reason for the better
	performance of DWRPM is that it uses the generalization
	ability of MLP and also captures correlation in daily
	rainfall values using convolutions, which is an important
	component of CNN. 
	This establishes the usefulness of the proposed method in
	prediction of rainfall in the Indian state of Rajasthan. 

\begin{table}[htb]
	\caption{Comparison of the proposed DWRPM with other
	deep learning methods, widely used for sequence-based
	prediction}
	\label{tab:Comparison}
\centering
	\begin{tabular}{lll}
	\hline
		\textbf{Method} & \textbf{MAE} & \textbf{RMSE} \\
	\hline
	MLP &  1.3137  & 2.7808 \\
	\hline
	1-DCNN & 0.8406 & 2.2894\\
	\hline
	LSTM & 0.8750 & 2.3095\\
	\hline
	DWRPM & 0.7765 & 2.1716\\
	\hline
\end{tabular}
\end{table}

\section{Conclusion and Future Work}
	\label{sec:Conclusion}
This paper has developed and presented a deep and wide
	network-based approach for the purpose of daily rainfall
	prediction in the Indian state of
	Rajasthan. Daily rainfall data of 71 years, from the year
	1957 to 2017, has been used in designing and validation of the
	proposed deep and wide rainfall prediction model. The results
	are promising and the model has generalization ability. Same
	model works well for forecasting rainfall in different
	atmospheric zones of Rajasthan.
	A comparison with the advance deep-learned-based models like
	MLP, LSTM and 1-DCNN is also presented. 
	The experimental analysis and comparison exhibits
	the importance of the proposed model for rainfall
	forecasting.
	While the model works well in prediction of light and
	moderate rainfall events, scope for improvement is there in
	prediction of heavy and very heavy rainfall events. Future
	work includes a comprehensive analysis of the applicability
	of the proposed model in different states of India.
	We shall also include more number of
	parameters and explore the ways to increase the forecasting
	accuracy for heavy and very heavy rainfall events.
	We also plan to estimate and predict
	the rainfall for longer duration of time.

\section{Acknowledgments}
This work is in collaboration with Water Resources, Government of
Rajasthan. We are thankful to Special Project Monitoring Unit,
National Hydrology Project, Water Resources Rajasthan Jaipur,
India for providing us the Rainfall data for this study.

\bibliographystyle{model2-names}
\bibliography{refs}

\begin{thebibliography}{80}
\expandafter\ifx\csname natexlab\endcsname\relax\def\natexlab#1{#1}\fi
\providecommand{\url}[1]{\texttt{#1}}
\providecommand{\href}[2]{#2}
\providecommand{\path}[1]{#1}
\providecommand{\DOIprefix}{doi:}
\providecommand{\ArXivprefix}{arXiv:}
\providecommand{\URLprefix}{URL: }
\providecommand{\Pubmedprefix}{pmid:}
\providecommand{\doi}[1]{\href{http://dx.doi.org/#1}{\path{#1}}}
\providecommand{\Pubmed}[1]{\href{pmid:#1}{\path{#1}}}
\providecommand{\bibinfo}[2]{#2}
\ifx\xfnm\relax \def\xfnm[#1]{\unskip,\space#1}\fi
\bibitem[{Abadi et~al.(2016)Abadi, Barham, Chen, Chen, Davis, Dean, Devin,
  Ghemawat, Irving, Isard et~al.}]{abadi2016tensorflow}
\bibinfo{author}{Abadi, M.}, \bibinfo{author}{Barham, P.},
  \bibinfo{author}{Chen, J.}, \bibinfo{author}{Chen, Z.},
  \bibinfo{author}{Davis, A.}, \bibinfo{author}{Dean, J.},
  \bibinfo{author}{Devin, M.}, \bibinfo{author}{Ghemawat, S.},
  \bibinfo{author}{Irving, G.}, \bibinfo{author}{Isard, M.}, et~al.,
  \bibinfo{year}{2016}.
\newblock \bibinfo{title}{Tensorflow: A system for large-scale machine
  learning}, in: \bibinfo{booktitle}{12th $\{$USENIX$\}$ symposium on operating
  systems design and implementation ($\{$OSDI$\}$ 16)}, pp.
  \bibinfo{pages}{265--283}.
\bibitem[{Acharya et~al.(2013)Acharya, Shrivastava, Panigrahi and
  Mohanty}]{Acharya13}
\bibinfo{author}{Acharya, N.}, \bibinfo{author}{Shrivastava, N.},
  \bibinfo{author}{Panigrahi, B.}, \bibinfo{author}{Mohanty, U.C.},
  \bibinfo{year}{2013}.
\newblock \bibinfo{title}{Development of an artificial neural network based
  multi-model ensemble to estimate the northeast monsoon rainfall over south
  peninsular india: An application of extreme learning machine}.
\newblock \bibinfo{journal}{Climate Dynamics} \bibinfo{volume}{43},
  \bibinfo{pages}{379--}.
\bibitem[{Agresti(2003)}]{agresti2003categorical}
\bibinfo{author}{Agresti, A.}, \bibinfo{year}{2003}.
\newblock \bibinfo{title}{Categorical data analysis}. volume
  \bibinfo{volume}{482}.
\newblock \bibinfo{publisher}{John Wiley \& Sons}.
\bibitem[{Al~Mamun et~al.(2018)Al~Mamun, bin Salleh and
  Noor}]{al2018estimation}
\bibinfo{author}{Al~Mamun, A.}, \bibinfo{author}{bin Salleh, M.N.},
  \bibinfo{author}{Noor, H.M.}, \bibinfo{year}{2018}.
\newblock \bibinfo{title}{Estimation of short-duration rainfall intensity from
  daily rainfall values in klang valley, malaysia}.
\newblock \bibinfo{journal}{Applied Water Science} \bibinfo{volume}{8},
  \bibinfo{pages}{203}.
\bibitem[{AlHassoun(2011)}]{alhassoun2011developing}
\bibinfo{author}{AlHassoun, S.A.}, \bibinfo{year}{2011}.
\newblock \bibinfo{title}{Developing an empirical formulae to estimate rainfall
  intensity in riyadh region}.
\newblock \bibinfo{journal}{Journal of King Saud University-Engineering
  Sciences} \bibinfo{volume}{23}, \bibinfo{pages}{81--88}.
\bibitem[{Awadallah et~al.(2017)Awadallah, Magdy, Helmy and
  Rashed}]{awadallah2017assessment}
\bibinfo{author}{Awadallah, A.G.}, \bibinfo{author}{Magdy, M.},
  \bibinfo{author}{Helmy, E.}, \bibinfo{author}{Rashed, E.},
  \bibinfo{year}{2017}.
\newblock \bibinfo{title}{Assessment of rainfall intensity equations enlisted
  in the egyptian code for designing potable water and sewage networks}.
\newblock \bibinfo{journal}{Advances in Meteorology} \bibinfo{volume}{2017}.
\bibitem[{Beheshti et~al.(2016)Beheshti, Firouzi, Shamsuddin, Zibarzani and
  Yusop}]{beheshti2016new}
\bibinfo{author}{Beheshti, Z.}, \bibinfo{author}{Firouzi, M.},
  \bibinfo{author}{Shamsuddin, S.M.}, \bibinfo{author}{Zibarzani, M.},
  \bibinfo{author}{Yusop, Z.}, \bibinfo{year}{2016}.
\newblock \bibinfo{title}{A new rainfall forecasting model using the capso
  algorithm and an artificial neural network}.
\newblock \bibinfo{journal}{Neural Computing and Applications}
  \bibinfo{volume}{27}, \bibinfo{pages}{2551--2565}.
\bibitem[{Benyon(1996)}]{benyon1996information}
\bibinfo{author}{Benyon, D.}, \bibinfo{year}{1996}.
\newblock \bibinfo{title}{Information and data modelling}.
\newblock \bibinfo{publisher}{McGraw-Hill Higher Education}.
\bibitem[{Bhatt et~al.(2013)Bhatt, Maskey, Babel, Uhlenbrook and
  Prasad}]{Bhatt13}
\bibinfo{author}{Bhatt, D.}, \bibinfo{author}{Maskey, S.},
  \bibinfo{author}{Babel, M.}, \bibinfo{author}{Uhlenbrook, S.},
  \bibinfo{author}{Prasad, K.}, \bibinfo{year}{2013}.
\newblock \bibinfo{title}{Climate trends and impacts on crop production in the
  koshi river basin of nepal}.
\newblock \bibinfo{journal}{Regional Environmental Change}
  \bibinfo{volume}{14}, \bibinfo{pages}{1291--1301}.
\bibitem[{Calvello et~al.(2008)Calvello, Cascini and
  Sorbino}]{calvello2008numerical}
\bibinfo{author}{Calvello, M.}, \bibinfo{author}{Cascini, L.},
  \bibinfo{author}{Sorbino, G.}, \bibinfo{year}{2008}.
\newblock \bibinfo{title}{A numerical procedure for predicting rainfall-induced
  movements of active landslides along pre-existing slip surfaces}.
\newblock \bibinfo{journal}{International Journal for Numerical and Analytical
  Methods in Geomechanics} \bibinfo{volume}{32}, \bibinfo{pages}{327--351}.
\bibitem[{Campling et~al.(2001)Campling, Gobin and Feyen}]{Campling98}
\bibinfo{author}{Campling, P.}, \bibinfo{author}{Gobin, A.},
  \bibinfo{author}{Feyen, J.}, \bibinfo{year}{2001}.
\newblock \bibinfo{title}{Temporal and spatial rainfall analysis across a humid
  tropical catchment}.
\newblock \bibinfo{journal}{Hydrological Processes} \bibinfo{volume}{15},
  \bibinfo{pages}{359--375}.
\bibitem[{Chakraborty et~al.(2018)Chakraborty, Saha, Sachdeva and
  Joshi}]{Chakraborty18}
\bibinfo{author}{Chakraborty, A.}, \bibinfo{author}{Saha, S.},
  \bibinfo{author}{Sachdeva, K.}, \bibinfo{author}{Joshi, P.},
  \bibinfo{year}{2018}.
\newblock \bibinfo{title}{Vulnerability of forests in the himalayan region to
  climate change impacts and anthropogenic disturbances: a systematic review}.
\newblock \bibinfo{journal}{Regional Environmental Change}
  \bibinfo{volume}{18}.
\bibitem[{Cheng et~al.(2016)Cheng, Koc, Harmsen, Shaked, Chandra, Aradhye,
  Anderson, Corrado, Chai, Ispir, Anil, Haque, Hong, Jain, Liu and
  Shah}]{Cheng16}
\bibinfo{author}{Cheng, H.T.}, \bibinfo{author}{Koc, L.},
  \bibinfo{author}{Harmsen, J.}, \bibinfo{author}{Shaked, T.},
  \bibinfo{author}{Chandra, T.}, \bibinfo{author}{Aradhye, H.},
  \bibinfo{author}{Anderson, G.}, \bibinfo{author}{Corrado, G.},
  \bibinfo{author}{Chai, W.}, \bibinfo{author}{Ispir, M.},
  \bibinfo{author}{Anil, R.}, \bibinfo{author}{Haque, Z.},
  \bibinfo{author}{Hong, L.}, \bibinfo{author}{Jain, V.}, \bibinfo{author}{Liu,
  X.}, \bibinfo{author}{Shah, H.}, \bibinfo{year}{2016}.
\newblock \bibinfo{title}{Wide \& deep learning for recommender systems}, in:
  \bibinfo{booktitle}{Proceedings of the 1st Workshop on Deep Learning for
  Recommender Systems}, p. \bibinfo{pages}{7–10}.
\bibitem[{Chollet et~al.(2015)}]{chollet2015keras}
\bibinfo{author}{Chollet, F.}, et~al., \bibinfo{year}{2015}.
\newblock \bibinfo{title}{Keras}.
\newblock \URLprefix \url{https://github.com/fchollet/keras}.
\bibitem[{Cramer et~al.(2017)Cramer, Kampouridis, Freitas and
  Alexandridis}]{cramer2017extensive}
\bibinfo{author}{Cramer, S.}, \bibinfo{author}{Kampouridis, M.},
  \bibinfo{author}{Freitas, A.A.}, \bibinfo{author}{Alexandridis, A.K.},
  \bibinfo{year}{2017}.
\newblock \bibinfo{title}{An extensive evaluation of seven machine learning
  methods for rainfall prediction in weather derivatives}.
\newblock \bibinfo{journal}{Expert Systems with Applications}
  \bibinfo{volume}{85}, \bibinfo{pages}{169--181}.
\bibitem[{Curci et~al.(2017)Curci, Lavecchia, Frustaci, Paolini, Pilati and
  Paganelli}]{curci2017assessing}
\bibinfo{author}{Curci, S.}, \bibinfo{author}{Lavecchia, C.},
  \bibinfo{author}{Frustaci, G.}, \bibinfo{author}{Paolini, R.},
  \bibinfo{author}{Pilati, S.}, \bibinfo{author}{Paganelli, C.},
  \bibinfo{year}{2017}.
\newblock \bibinfo{title}{Assessing measurement uncertainty in meteorology in
  urban environments}.
\newblock \bibinfo{journal}{Measurement Science and Technology}
  \bibinfo{volume}{28}, \bibinfo{pages}{104002}.
\bibitem[{Dabhi and Chaudhary(2014)}]{Dabhi14}
\bibinfo{author}{Dabhi, V.}, \bibinfo{author}{Chaudhary, S.},
  \bibinfo{year}{2014}.
\newblock \bibinfo{title}{Hybrid wavelet-postfix-gp model for rainfall
  prediction of anand region of india}.
\newblock \bibinfo{journal}{Advances in Artificial Intelligence}
  \bibinfo{volume}{2014}, \bibinfo{pages}{11}.
\bibitem[{Darji et~al.(2015)Darji, Dabhi and Prajapati}]{Darji15}
\bibinfo{author}{Darji, M.}, \bibinfo{author}{Dabhi, V.},
  \bibinfo{author}{Prajapati, H.}, \bibinfo{year}{2015}.
\newblock \bibinfo{title}{Rainfall forecasting using neural network: A survey},
  in: \bibinfo{booktitle}{International Conference on Advances in Computer
  Engineering and Applications}, pp. \bibinfo{pages}{706--713}.
\bibitem[{Dubey(2015a)}]{Dubey15}
\bibinfo{author}{Dubey, A.D.}, \bibinfo{year}{2015}a.
\newblock \bibinfo{title}{Article: Artificial neural network models for
  rainfall prediction in pondicherry}.
\newblock \bibinfo{journal}{International Journal of Computer Applications}
  \bibinfo{volume}{120}, \bibinfo{pages}{30--35}.
\bibitem[{Dubey(2015b)}]{dubey2015artificial}
\bibinfo{author}{Dubey, A.D.}, \bibinfo{year}{2015}b.
\newblock \bibinfo{title}{Artificial neural network models for rainfall
  prediction in pondicherry}.
\newblock \bibinfo{journal}{International Journal of Computer Applications}
  \bibinfo{volume}{120}.
\bibitem[{Ducrocq et~al.(2002)Ducrocq, Ricard, Lafore and
  Orain}]{ducrocq2002storm}
\bibinfo{author}{Ducrocq, V.}, \bibinfo{author}{Ricard, D.},
  \bibinfo{author}{Lafore, J.P.}, \bibinfo{author}{Orain, F.},
  \bibinfo{year}{2002}.
\newblock \bibinfo{title}{Storm-scale numerical rainfall prediction for five
  precipitating events over france: On the importance of the initial humidity
  field}.
\newblock \bibinfo{journal}{Weather and Forecasting} \bibinfo{volume}{17},
  \bibinfo{pages}{1236--1256}.
\bibitem[{Dutta et~al.(2013)Dutta, Kundu and Patel}]{dutta2013predicting}
\bibinfo{author}{Dutta, D.}, \bibinfo{author}{Kundu, A.},
  \bibinfo{author}{Patel, N.}, \bibinfo{year}{2013}.
\newblock \bibinfo{title}{Predicting agricultural drought in eastern rajasthan
  of india using ndvi and standardized precipitation index}.
\newblock \bibinfo{journal}{Geocarto International} \bibinfo{volume}{28},
  \bibinfo{pages}{192--209}.
\bibitem[{Esteves et~al.(2019)Esteves, de~Souza~Rolim and
  Ferraudo}]{esteves2019rainfall}
\bibinfo{author}{Esteves, J.T.}, \bibinfo{author}{de~Souza~Rolim, G.},
  \bibinfo{author}{Ferraudo, A.S.}, \bibinfo{year}{2019}.
\newblock \bibinfo{title}{Rainfall prediction methodology with binary
  multilayer perceptron neural networks}.
\newblock \bibinfo{journal}{Climate Dynamics} \bibinfo{volume}{52},
  \bibinfo{pages}{2319--2331}.
\bibitem[{Gadgil(2003)}]{gadgil2003indian}
\bibinfo{author}{Gadgil, S.}, \bibinfo{year}{2003}.
\newblock \bibinfo{title}{The indian monsoon and its variability}.
\newblock \bibinfo{journal}{Annual Review of Earth and Planetary Sciences}
  \bibinfo{volume}{31}, \bibinfo{pages}{429--467}.
\bibitem[{Gadgil et~al.(2005)Gadgil, Rajeevan and
  Nanjundiah}]{gadgil2005monsoon}
\bibinfo{author}{Gadgil, S.}, \bibinfo{author}{Rajeevan, M.},
  \bibinfo{author}{Nanjundiah, R.}, \bibinfo{year}{2005}.
\newblock \bibinfo{title}{Monsoon prediction—why yet another failure}.
\newblock \bibinfo{journal}{Curr. Sci} \bibinfo{volume}{88},
  \bibinfo{pages}{1389--1400}.
\bibitem[{Gadgil et~al.(2002)Gadgil, Srinivasan, Nanjundiah, Kumar, Munot and
  Kumar}]{gadgil2002forecasting}
\bibinfo{author}{Gadgil, S.}, \bibinfo{author}{Srinivasan, J.},
  \bibinfo{author}{Nanjundiah, R.S.}, \bibinfo{author}{Kumar, K.K.},
  \bibinfo{author}{Munot, A.}, \bibinfo{author}{Kumar, K.R.},
  \bibinfo{year}{2002}.
\newblock \bibinfo{title}{On forecasting the indian summer monsoon: the
  intriguing season of 2002}.
\newblock \bibinfo{journal}{Current Science} \bibinfo{volume}{83},
  \bibinfo{pages}{394--403}.
\bibitem[{Glorot and Bengio(2010)}]{Glorot10}
\bibinfo{author}{Glorot, X.}, \bibinfo{author}{Bengio, Y.},
  \bibinfo{year}{2010}.
\newblock \bibinfo{title}{Understanding the difficulty of training deep
  feedforward neural networks}, in: \bibinfo{booktitle}{In Proceedings of the
  International Conference on Artificial Intelligence and Statistics
  (AISTATS’10). Society for Artificial Intelligence and Statistics}, pp.
  \bibinfo{pages}{249--256}.
\bibitem[{Gope et~al.(2016)Gope, Sarkar, Mitra and Ghosh}]{gope2016early}
\bibinfo{author}{Gope, S.}, \bibinfo{author}{Sarkar, S.},
  \bibinfo{author}{Mitra, P.}, \bibinfo{author}{Ghosh, S.},
  \bibinfo{year}{2016}.
\newblock \bibinfo{title}{Early prediction of extreme rainfall events: a deep
  learning approach}, in: \bibinfo{booktitle}{Industrial Conference on Data
  Mining}, \bibinfo{organization}{Springer}. pp. \bibinfo{pages}{154--167}.
\bibitem[{Goyal(2004)}]{goyal2004sensitivity}
\bibinfo{author}{Goyal, R.}, \bibinfo{year}{2004}.
\newblock \bibinfo{title}{Sensitivity of evapotranspiration to global warming:
  a case study of arid zone of rajasthan (india)}.
\newblock \bibinfo{journal}{Agricultural water management}
  \bibinfo{volume}{69}, \bibinfo{pages}{1--11}.
\bibitem[{Gu et~al.(2018)Gu, Wang, Kuen, Ma, Shahroudy, Shuai, Liu, Wang, Wang,
  Cai et~al.}]{gu2018recent}
\bibinfo{author}{Gu, J.}, \bibinfo{author}{Wang, Z.}, \bibinfo{author}{Kuen,
  J.}, \bibinfo{author}{Ma, L.}, \bibinfo{author}{Shahroudy, A.},
  \bibinfo{author}{Shuai, B.}, \bibinfo{author}{Liu, T.},
  \bibinfo{author}{Wang, X.}, \bibinfo{author}{Wang, G.}, \bibinfo{author}{Cai,
  J.}, et~al., \bibinfo{year}{2018}.
\newblock \bibinfo{title}{Recent advances in convolutional neural networks}.
\newblock \bibinfo{journal}{Pattern Recognition} \bibinfo{volume}{77},
  \bibinfo{pages}{354--377}.
\bibitem[{Gulli and Pal(2017)}]{gulli2017deep}
\bibinfo{author}{Gulli, A.}, \bibinfo{author}{Pal, S.}, \bibinfo{year}{2017}.
\newblock \bibinfo{title}{Deep learning with Keras}.
\newblock \bibinfo{publisher}{Packt Publishing Ltd}.
\bibitem[{Halbe et~al.(2013)Halbe, Pahl-Wostl, Sendzimir and
  Adamowski}]{Halbe13}
\bibinfo{author}{Halbe, J.}, \bibinfo{author}{Pahl-Wostl, C.},
  \bibinfo{author}{Sendzimir, J.}, \bibinfo{author}{Adamowski, J.},
  \bibinfo{year}{2013}.
\newblock \bibinfo{title}{Towards adaptive and integrated management paradigms
  to meet the challenges of water governance}.
\newblock \bibinfo{journal}{Water science and technology : a journal of the
  International Association on Water Pollution Research} \bibinfo{volume}{67},
  \bibinfo{pages}{2651--60}.
\bibitem[{Hardwinarto et~al.(2015)Hardwinarto, Aipassa
  et~al.}]{hardwinarto2015rainfall}
\bibinfo{author}{Hardwinarto, S.}, \bibinfo{author}{Aipassa, M.}, et~al.,
  \bibinfo{year}{2015}.
\newblock \bibinfo{title}{Rainfall monthly prediction based on artificial
  neural network: a case study in tenggarong station, east
  kalimantan-indonesia}.
\newblock \bibinfo{journal}{Procedia Computer Science} \bibinfo{volume}{59},
  \bibinfo{pages}{142--151}.
\bibitem[{He et~al.(2015a)He, Zhang, Ren and Sun}]{He15}
\bibinfo{author}{He, K.}, \bibinfo{author}{Zhang, X.}, \bibinfo{author}{Ren,
  S.}, \bibinfo{author}{Sun, J.}, \bibinfo{year}{2015}a.
\newblock \bibinfo{title}{Delving deep into rectifiers: Surpassing human-level
  performance on imagenet classification}, in: \bibinfo{booktitle}{Proceedings
  of the 2015 IEEE International Conference on Computer Vision (ICCV)}, p.
  \bibinfo{pages}{1026–1034}.
\bibitem[{He et~al.(2015b)He, Zhang, Ren and Sun}]{he2015delving}
\bibinfo{author}{He, K.}, \bibinfo{author}{Zhang, X.}, \bibinfo{author}{Ren,
  S.}, \bibinfo{author}{Sun, J.}, \bibinfo{year}{2015}b.
\newblock \bibinfo{title}{Delving deep into rectifiers: Surpassing human-level
  performance on imagenet classification}, in: \bibinfo{booktitle}{Proceedings
  of the IEEE international conference on computer vision}, pp.
  \bibinfo{pages}{1026--1034}.
\bibitem[{Hern{\'a}ndez et~al.(2016)Hern{\'a}ndez, Sanchez-Anguix, Julian,
  Palanca and Duque}]{hernandez2016:rainfall}
\bibinfo{author}{Hern{\'a}ndez, E.}, \bibinfo{author}{Sanchez-Anguix, V.},
  \bibinfo{author}{Julian, V.}, \bibinfo{author}{Palanca, J.},
  \bibinfo{author}{Duque, N.}, \bibinfo{year}{2016}.
\newblock \bibinfo{title}{Rainfall prediction: A deep learning approach}, in:
  \bibinfo{booktitle}{International Conference on Hybrid Artificial
  Intelligence Systems}, \bibinfo{organization}{Springer}. pp.
  \bibinfo{pages}{151--162}.
\bibitem[{Hern{\'a}ndez and Stolfo(1998)}]{hernandez1998real}
\bibinfo{author}{Hern{\'a}ndez, M.A.}, \bibinfo{author}{Stolfo, S.J.},
  \bibinfo{year}{1998}.
\newblock \bibinfo{title}{Real-world data is dirty: Data cleansing and the
  merge/purge problem}.
\newblock \bibinfo{journal}{Data mining and knowledge discovery}
  \bibinfo{volume}{2}, \bibinfo{pages}{9--37}.
\bibitem[{Karunasagar et~al.(2017)Karunasagar, Rajeevan, Rao and
  Mitra}]{Karunasagar17}
\bibinfo{author}{Karunasagar, S.}, \bibinfo{author}{Rajeevan, M.},
  \bibinfo{author}{Rao, S.}, \bibinfo{author}{Mitra, A.}, \bibinfo{year}{2017}.
\newblock \bibinfo{title}{Prediction skill of rainstorm events over india in
  the tigge weather prediction models}.
\newblock \bibinfo{journal}{Atmospheric Research} \bibinfo{volume}{198}.
\bibitem[{Kashiwao et~al.(2017)Kashiwao, Nakayama, Ando, Ikeda, Lee and
  Bahadori}]{Kashiwao17}
\bibinfo{author}{Kashiwao, T.}, \bibinfo{author}{Nakayama, K.},
  \bibinfo{author}{Ando, S.}, \bibinfo{author}{Ikeda, K.},
  \bibinfo{author}{Lee, M.}, \bibinfo{author}{Bahadori, A.},
  \bibinfo{year}{2017}.
\newblock \bibinfo{title}{A neural network-based local rainfall prediction
  system using meteorological data on the internet: A case study using data
  from the japan meteorological agency}.
\newblock \bibinfo{journal}{Applied Soft Computing} \bibinfo{volume}{56},
  \bibinfo{pages}{317 -- 330}.
\bibitem[{Kim et~al.(2020)Kim, Lee and Kim}]{Kim20}
\bibinfo{author}{Kim, M.}, \bibinfo{author}{Lee, S.}, \bibinfo{author}{Kim,
  J.}, \bibinfo{year}{2020}.
\newblock \bibinfo{title}{A wide deep learning sharing input data for
  regression analysis}, in: \bibinfo{booktitle}{2020 IEEE International
  Conference on Big Data and Smart Computing (BigComp)}, pp.
  \bibinfo{pages}{8--12}.
\bibitem[{Kingma and Ba(2015)}]{kingma14}
\bibinfo{author}{Kingma, D.P.}, \bibinfo{author}{Ba, J.}, \bibinfo{year}{2015}.
\newblock \bibinfo{title}{Adam: A method for stochastic optimization}, in:
  \bibinfo{booktitle}{Proceedings of the 3rd International Conference for
  Learning Representations (ICLR)}.
\bibitem[{Ko et~al.(2020)Ko, Jeong, Lee and Kim}]{ko2020development}
\bibinfo{author}{Ko, C.M.}, \bibinfo{author}{Jeong, Y.Y.},
  \bibinfo{author}{Lee, Y.M.}, \bibinfo{author}{Kim, B.S.},
  \bibinfo{year}{2020}.
\newblock \bibinfo{title}{The development of a quantitative precipitation
  forecast correction technique based on machine learning for hydrological
  applications}.
\newblock \bibinfo{journal}{Atmosphere} \bibinfo{volume}{11},
  \bibinfo{pages}{111}.
\bibitem[{Kumar et~al.(2006)Kumar, Rajagopalan, Hoerling, Bates and
  Cane}]{kumar2006unraveling}
\bibinfo{author}{Kumar, K.K.}, \bibinfo{author}{Rajagopalan, B.},
  \bibinfo{author}{Hoerling, M.}, \bibinfo{author}{Bates, G.},
  \bibinfo{author}{Cane, M.}, \bibinfo{year}{2006}.
\newblock \bibinfo{title}{Unraveling the mystery of indian monsoon failure
  during el ni{\~n}o}.
\newblock \bibinfo{journal}{Science} \bibinfo{volume}{314},
  \bibinfo{pages}{115--119}.
\bibitem[{Kumar et~al.(2018)Kumar, Kulkarni and Gupta}]{Kumar18}
\bibinfo{author}{Kumar, V.}, \bibinfo{author}{Kulkarni, A.},
  \bibinfo{author}{Gupta, A.}, \bibinfo{year}{2018}.
\newblock \bibinfo{title}{Assessment of snow-glacier melt and rainfall
  contribution to stream runoff in baspa basin, indian himalaya}.
\newblock \bibinfo{journal}{Environmental Monitoring and Assessment}
  \bibinfo{volume}{190}.
\bibitem[{Le and Vo(2020)}]{le2020livelihood}
\bibinfo{author}{Le, S.T.}, \bibinfo{author}{Vo, C.D.}, \bibinfo{year}{2020}.
\newblock \bibinfo{title}{The livelihood adaptability of households under the
  impact of climate change in the mekong delta}.
\newblock \bibinfo{journal}{Journal of Agribusiness in Developing and Emerging
  Economies} .
\bibitem[{Li and Shao(2010)}]{li2010improved}
\bibinfo{author}{Li, M.}, \bibinfo{author}{Shao, Q.}, \bibinfo{year}{2010}.
\newblock \bibinfo{title}{An improved statistical approach to merge satellite
  rainfall estimates and raingauge data}.
\newblock \bibinfo{journal}{Journal of Hydrology} \bibinfo{volume}{385},
  \bibinfo{pages}{51--64}.
\bibitem[{Liu et~al.(2019)Liu, Zou, Liu and Linge}]{Liu19}
\bibinfo{author}{Liu, Q.}, \bibinfo{author}{Zou, Y.}, \bibinfo{author}{Liu,
  X.}, \bibinfo{author}{Linge, N.}, \bibinfo{year}{2019}.
\newblock \bibinfo{title}{A survey on rainfall forecasting using artificial
  neural network}.
\newblock \bibinfo{journal}{International Journal of Embedded Systems (IJES)}
  \bibinfo{volume}{11}, \bibinfo{pages}{240--249}.
\bibitem[{Mishra et~al.()Mishra, Nagaraju and Meer}]{mishraexamination}
\bibinfo{author}{Mishra, A.K.}, \bibinfo{author}{Nagaraju, V.},
  \bibinfo{author}{Meer, M.S.}, .
\newblock \bibinfo{title}{Examination of heavy flooding over the desert state
  of rajasthan in india}.
\newblock \bibinfo{journal}{Weather} .
\bibitem[{Montanari and Grossi(2008)}]{montanari2008estimating}
\bibinfo{author}{Montanari, A.}, \bibinfo{author}{Grossi, G.},
  \bibinfo{year}{2008}.
\newblock \bibinfo{title}{Estimating the uncertainty of hydrological forecasts:
  A statistical approach}.
\newblock \bibinfo{journal}{Water Resources Research} \bibinfo{volume}{44}.
\bibitem[{Namitha et~al.(2015)Namitha, Jayapriya and
  Kumar}]{namitha2015rainfall}
\bibinfo{author}{Namitha, K.}, \bibinfo{author}{Jayapriya, A.},
  \bibinfo{author}{Kumar, G.S.}, \bibinfo{year}{2015}.
\newblock \bibinfo{title}{Rainfall prediction using artificial neural network
  on map-reduce framework}, in: \bibinfo{booktitle}{Proceedings of the Third
  International Symposium on Women in Computing and Informatics}, pp.
  \bibinfo{pages}{492--495}.
\bibitem[{Nayak et~al.(2013)Nayak, Mahapatra and Mishra}]{Nayak13}
\bibinfo{author}{Nayak, D.R.}, \bibinfo{author}{Mahapatra, A.},
  \bibinfo{author}{Mishra, P.}, \bibinfo{year}{2013}.
\newblock \bibinfo{title}{A survey on rainfall prediction using artificial
  neural network}.
\newblock \bibinfo{journal}{International Journal of Computer Applications}
  \bibinfo{volume}{72}, \bibinfo{pages}{32--40}.
\bibitem[{Ni et~al.(2020)Ni, Wang, Singh, Wu, Wang, Tao and
  Zhang}]{ni2020streamflow}
\bibinfo{author}{Ni, L.}, \bibinfo{author}{Wang, D.}, \bibinfo{author}{Singh,
  V.P.}, \bibinfo{author}{Wu, J.}, \bibinfo{author}{Wang, Y.},
  \bibinfo{author}{Tao, Y.}, \bibinfo{author}{Zhang, J.}, \bibinfo{year}{2020}.
\newblock \bibinfo{title}{Streamflow and rainfall forecasting by two long
  short-term memory-based models}.
\newblock \bibinfo{journal}{Journal of Hydrology} \bibinfo{volume}{583},
  \bibinfo{pages}{124296}.
\bibitem[{Ortiz-Garcia et~al.(2014)Ortiz-Garcia, Salcedo-Sanz and
  Casanova}]{OrtizGarcia14}
\bibinfo{author}{Ortiz-Garcia, E.}, \bibinfo{author}{Salcedo-Sanz, S.},
  \bibinfo{author}{Casanova, C.}, \bibinfo{year}{2014}.
\newblock \bibinfo{title}{Accurate precipitation prediction with support vector
  classifiers: A study including novel predictive variables and observational
  data}.
\newblock \bibinfo{journal}{Atmospheric Research} \bibinfo{volume}{139},
  \bibinfo{pages}{128–136}.
\bibitem[{Pal and Mitra(1992)}]{pal1992multilayer}
\bibinfo{author}{Pal, S.K.}, \bibinfo{author}{Mitra, S.}, \bibinfo{year}{1992}.
\newblock \bibinfo{title}{Multilayer perceptron, fuzzy sets, classifiaction} .
\bibitem[{Pangaluru et~al.(2015)Pangaluru, Jyothi, Basha, Rao, Rajeevan,
  Velicogna, Sutterley and Sutterley}]{Pangaluru15}
\bibinfo{author}{Pangaluru, K.}, \bibinfo{author}{Jyothi, S.},
  \bibinfo{author}{Basha, G.}, \bibinfo{author}{Rao, S.},
  \bibinfo{author}{Rajeevan, M.}, \bibinfo{author}{Velicogna, I.},
  \bibinfo{author}{Sutterley, T.}, \bibinfo{author}{Sutterley, C.},
  \bibinfo{year}{2015}.
\newblock \bibinfo{title}{Precipitation climatology over india: validation with
  observations and reanalysis datasets and spatial trends}.
\newblock \bibinfo{journal}{Climate Dynamics} \bibinfo{volume}{46}.
\bibitem[{Parthasarathy et~al.(1992)Parthasarathy, Kumar and
  Kothawale}]{parthasarathy1992indian}
\bibinfo{author}{Parthasarathy, B.}, \bibinfo{author}{Kumar, K.R.},
  \bibinfo{author}{Kothawale, D.}, \bibinfo{year}{1992}.
\newblock \bibinfo{title}{Indian summer monsoon rainfall indices: 1871-1990} .
\bibitem[{Parthasarathy et~al.(1994)Parthasarathy, Munot and
  Kothawale}]{parthasarathy1994all}
\bibinfo{author}{Parthasarathy, B.}, \bibinfo{author}{Munot, A.},
  \bibinfo{author}{Kothawale, D.}, \bibinfo{year}{1994}.
\newblock \bibinfo{title}{All-india monthly and seasonal rainfall series:
  1871--1993}.
\newblock \bibinfo{journal}{Theoretical and Applied Climatology}
  \bibinfo{volume}{49}, \bibinfo{pages}{217--224}.
\bibitem[{Pham et~al.(2020)Pham, Le, Le, Bui, Le, Ly and
  Prakash}]{pham2020development}
\bibinfo{author}{Pham, B.T.}, \bibinfo{author}{Le, L.M.}, \bibinfo{author}{Le,
  T.T.}, \bibinfo{author}{Bui, K.T.T.}, \bibinfo{author}{Le, V.M.},
  \bibinfo{author}{Ly, H.B.}, \bibinfo{author}{Prakash, I.},
  \bibinfo{year}{2020}.
\newblock \bibinfo{title}{Development of advanced artificial intelligence
  models for daily rainfall prediction}.
\newblock \bibinfo{journal}{Atmospheric Research} \bibinfo{volume}{237},
  \bibinfo{pages}{104845}.
\bibitem[{Preethi et~al.(2011)Preethi, Revadekar and
  Kripalani}]{preethi2011anomalous}
\bibinfo{author}{Preethi, B.}, \bibinfo{author}{Revadekar, J.},
  \bibinfo{author}{Kripalani, R.}, \bibinfo{year}{2011}.
\newblock \bibinfo{title}{Anomalous behaviour of the indian summer monsoon
  2009}.
\newblock \bibinfo{journal}{Journal of earth system science}
  \bibinfo{volume}{120}, \bibinfo{pages}{783--794}.
\bibitem[{Ren et~al.(2020)Ren, Meng, Wang, Lu and Yang}]{Ren20}
\bibinfo{author}{Ren, L.}, \bibinfo{author}{Meng, Z.}, \bibinfo{author}{Wang,
  X.}, \bibinfo{author}{Lu, R.}, \bibinfo{author}{Yang, L.},
  \bibinfo{year}{2020}.
\newblock \bibinfo{title}{A wide-deep-sequence model-based quality prediction
  method in industrial process analysis}.
\newblock \bibinfo{journal}{IEEE Transactions on Neural Networks and Learning
  Systems} \bibinfo{volume}{PP}, \bibinfo{pages}{1--11}.
\bibitem[{Saha et~al.(2020)Saha, Santara, Mitra, Chakraborty and
  Nanjundiah}]{saha2020prediction}
\bibinfo{author}{Saha, M.}, \bibinfo{author}{Santara, A.},
  \bibinfo{author}{Mitra, P.}, \bibinfo{author}{Chakraborty, A.},
  \bibinfo{author}{Nanjundiah, R.S.}, \bibinfo{year}{2020}.
\newblock \bibinfo{title}{Prediction of the indian summer monsoon using a
  stacked autoencoder and ensemble regression model}.
\newblock \bibinfo{journal}{International Journal of Forecasting} .
\bibitem[{Sahai et~al.(2000)Sahai, Soman and Satyan}]{Sahai00}
\bibinfo{author}{Sahai, A.}, \bibinfo{author}{Soman, M.},
  \bibinfo{author}{Satyan, V.}, \bibinfo{year}{2000}.
\newblock \bibinfo{title}{All india summer monsoon rainfall prediction using an
  artificial neural network}.
\newblock \bibinfo{journal}{Climate Dynamics} \bibinfo{volume}{16},
  \bibinfo{pages}{291--302}.
\bibitem[{Samantaray et~al.(2020)Samantaray, Tripathy, Sahoo and
  Ghose}]{samantaray2020rainfall}
\bibinfo{author}{Samantaray, S.}, \bibinfo{author}{Tripathy, O.},
  \bibinfo{author}{Sahoo, A.}, \bibinfo{author}{Ghose, D.K.},
  \bibinfo{year}{2020}.
\newblock \bibinfo{title}{Rainfall forecasting through ann and svm in bolangir
  watershed, india}, in: \bibinfo{booktitle}{Smart Intelligent Computing and
  Applications}. \bibinfo{publisher}{Springer}, pp. \bibinfo{pages}{767--774}.
\bibitem[{Sankaranarayanan et~al.(2019)Sankaranarayanan, Prabhakar, Satish,
  Jain, Ramprasad and Krishnan}]{Sankaranarayanan19}
\bibinfo{author}{Sankaranarayanan, S.}, \bibinfo{author}{Prabhakar, M.},
  \bibinfo{author}{Satish, S.}, \bibinfo{author}{Jain, P.},
  \bibinfo{author}{Ramprasad, A.}, \bibinfo{author}{Krishnan, A.},
  \bibinfo{year}{2019}.
\newblock \bibinfo{title}{{Flood prediction based on weather parameters using
  deep learning}}.
\newblock \bibinfo{journal}{Journal of Water and Climate Change} .
\bibitem[{Sapsford and Jupp(1996)}]{sapsford1996data}
\bibinfo{author}{Sapsford, R.}, \bibinfo{author}{Jupp, V.},
  \bibinfo{year}{1996}.
\newblock \bibinfo{title}{Data collection and analysis}.
\newblock \bibinfo{publisher}{Sage}.
\bibitem[{Shah et~al.(2018)Shah, Garg, Sisodiya, Dube and
  Sharma}]{shah2018rainfall}
\bibinfo{author}{Shah, U.}, \bibinfo{author}{Garg, S.},
  \bibinfo{author}{Sisodiya, N.}, \bibinfo{author}{Dube, N.},
  \bibinfo{author}{Sharma, S.}, \bibinfo{year}{2018}.
\newblock \bibinfo{title}{Rainfall prediction: Accuracy enhancement using
  machine learning and forecasting techniques}, in: \bibinfo{booktitle}{Fifth
  International Conference on Parallel, Distributed and Grid Computing (PDGC)},
  \bibinfo{organization}{IEEE}. pp. \bibinfo{pages}{776--782}.
\bibitem[{Shanker et~al.(1996)Shanker, Hu and Hung}]{shanker1996effect}
\bibinfo{author}{Shanker, M.}, \bibinfo{author}{Hu, M.Y.},
  \bibinfo{author}{Hung, M.S.}, \bibinfo{year}{1996}.
\newblock \bibinfo{title}{Effect of data standardization on neural network
  training}.
\newblock \bibinfo{journal}{Omega} \bibinfo{volume}{24},
  \bibinfo{pages}{385--397}.
\bibitem[{Singh et~al.(2012a)Singh, Kulkarni, Mohanty, Kar, Robertson and
  Mishra}]{singh2012prediction}
\bibinfo{author}{Singh, A.}, \bibinfo{author}{Kulkarni, M.A.},
  \bibinfo{author}{Mohanty, U.}, \bibinfo{author}{Kar, S.},
  \bibinfo{author}{Robertson, A.W.}, \bibinfo{author}{Mishra, G.},
  \bibinfo{year}{2012}a.
\newblock \bibinfo{title}{Prediction of indian summer monsoon rainfall (ismr)
  using canonical correlation analysis of global circulation model products}.
\newblock \bibinfo{journal}{Meteorological Applications} \bibinfo{volume}{19},
  \bibinfo{pages}{179--188}.
\bibitem[{Singh et~al.(2012b)Singh, Rajpurohit, Vasishth and
  Singh}]{singh2012probability}
\bibinfo{author}{Singh, B.}, \bibinfo{author}{Rajpurohit, D.},
  \bibinfo{author}{Vasishth, A.}, \bibinfo{author}{Singh, J.},
  \bibinfo{year}{2012}b.
\newblock \bibinfo{title}{Probability analysis for estimation of annual one day
  maximum rainfall of jhalarapatan area of rajasthan, india}.
\newblock \bibinfo{journal}{Plant Archives} \bibinfo{volume}{12},
  \bibinfo{pages}{1093--1100}.
\bibitem[{Singh(2017)}]{Singh17}
\bibinfo{author}{Singh, P.}, \bibinfo{year}{2017}.
\newblock \bibinfo{title}{Indian summer monsoon rainfall (ismr) forecasting
  using time series data: A fuzzy-entropy-neuro based expert system}.
\newblock \bibinfo{journal}{Geoscience Frontiers} \bibinfo{volume}{9}.
\bibitem[{Singh and Borah(2013)}]{Singh13}
\bibinfo{author}{Singh, P.}, \bibinfo{author}{Borah, B.}, \bibinfo{year}{2013}.
\newblock \bibinfo{title}{Indian summer monsoon rainfall prediction using
  artificial neural network}.
\newblock \bibinfo{journal}{Stochastic Environmental Research and Risk
  Assessment} \bibinfo{volume}{27}.
\bibitem[{Srivastava et~al.(2014)Srivastava, Hinton, Krizhevsky, Sutskever and
  Salakhutdinov}]{srivastava2014dropout}
\bibinfo{author}{Srivastava, N.}, \bibinfo{author}{Hinton, G.},
  \bibinfo{author}{Krizhevsky, A.}, \bibinfo{author}{Sutskever, I.},
  \bibinfo{author}{Salakhutdinov, R.}, \bibinfo{year}{2014}.
\newblock \bibinfo{title}{Dropout: a simple way to prevent neural networks from
  overfitting}.
\newblock \bibinfo{journal}{The journal of machine learning research}
  \bibinfo{volume}{15}, \bibinfo{pages}{1929--1958}.
\bibitem[{Upadhyaya(2014)}]{Upadhyaya14}
\bibinfo{author}{Upadhyaya, H.}, \bibinfo{year}{2014}.
\newblock \bibinfo{title}{Vulnerability and Adaptation to Climate Change in the
  Context of Water Resource with Reference to Rajasthan}.
\newblock Ph.D. thesis. The IIS University.
\bibitem[{Vaes et~al.(2001)Vaes, Willems and Berlamont}]{vaes2001rainfall}
\bibinfo{author}{Vaes, G.}, \bibinfo{author}{Willems, P.},
  \bibinfo{author}{Berlamont, J.}, \bibinfo{year}{2001}.
\newblock \bibinfo{title}{Rainfall input requirements for hydrological
  calculations}.
\newblock \bibinfo{journal}{Urban Water} \bibinfo{volume}{3},
  \bibinfo{pages}{107--112}.
\bibitem[{Van et~al.(2020)Van, Le, Thanh, Dang, Loc and Anh}]{Van20}
\bibinfo{author}{Van, S.P.}, \bibinfo{author}{Le, H.M.},
  \bibinfo{author}{Thanh, D.V.}, \bibinfo{author}{Dang, T.D.},
  \bibinfo{author}{Loc, H.H.}, \bibinfo{author}{Anh, D.T.},
  \bibinfo{year}{2020}.
\newblock \bibinfo{title}{{Deep learning convolutional neural network in
  rainfall–runoff modelling}}.
\newblock \bibinfo{journal}{Journal of Hydroinformatics} \bibinfo{volume}{22},
  \bibinfo{pages}{541--561}.
\bibitem[{Weller and Romney(1988)}]{weller1988systematic}
\bibinfo{author}{Weller, S.C.}, \bibinfo{author}{Romney, A.K.},
  \bibinfo{year}{1988}.
\newblock \bibinfo{title}{Systematic data collection}.
  volume~\bibinfo{volume}{10}.
\newblock \bibinfo{publisher}{Sage publications}.
\bibitem[{Xingjian et~al.(2015)Xingjian, Chen, Wang, Yeung, Wong and
  Woo}]{xingjian2015convolutional}
\bibinfo{author}{Xingjian, S.}, \bibinfo{author}{Chen, Z.},
  \bibinfo{author}{Wang, H.}, \bibinfo{author}{Yeung, D.Y.},
  \bibinfo{author}{Wong, W.K.}, \bibinfo{author}{Woo, W.c.},
  \bibinfo{year}{2015}.
\newblock \bibinfo{title}{Convolutional lstm network: A machine learning
  approach for precipitation nowcasting}, in: \bibinfo{booktitle}{Advances in
  neural information processing systems}, pp. \bibinfo{pages}{802--810}.
\bibitem[{Zaman(2018)}]{Zaman18}
\bibinfo{author}{Zaman, Y.}, \bibinfo{year}{2018}.
\newblock \bibinfo{title}{Machine learning model on rainfall - a predicted
  approach for bangladesh}, in: \bibinfo{booktitle}{MS dissertation, United
  International University, Dhaka, Bangladesh}.
\bibitem[{Zhang et~al.(2020a)Zhang, Cao and Li}]{zhang2020surface}
\bibinfo{author}{Zhang, P.}, \bibinfo{author}{Cao, W.}, \bibinfo{author}{Li,
  W.}, \bibinfo{year}{2020}a.
\newblock \bibinfo{title}{Surface and high-altitude combined rainfall
  forecasting using convolutional neural network}.
\newblock \bibinfo{journal}{Peer-to-Peer Networking and Applications} ,
  \bibinfo{pages}{1--13}.
\bibitem[{Zhang et~al.(2020b)Zhang, Mohanty, Parida and Pani}]{Zhang20}
\bibinfo{author}{Zhang, X.}, \bibinfo{author}{Mohanty, S.},
  \bibinfo{author}{Parida, A.}, \bibinfo{author}{Pani, D.S.},
  \bibinfo{year}{2020}b.
\newblock \bibinfo{title}{Annual and non-monsoon rainfall prediction modelling
  using svr-mlp: An empirical study from odisha}.
\newblock \bibinfo{journal}{IEEE Access} \bibinfo{volume}{8},
  \bibinfo{pages}{1--1}.

\end{thebibliography}

\end{document}